\documentclass[pdflatex,sn-mathphys-num]{sn-jnl}

\usepackage{graphicx}%
\usepackage{multirow}%
\usepackage{amsmath,amssymb,amsfonts}%
\usepackage{amsthm}%
\usepackage{mathrsfs}%
\usepackage[title]{appendix}%
\usepackage{xcolor}%
\usepackage{textcomp}%
\usepackage{manyfoot}%
\usepackage{booktabs}%
\usepackage{algorithm}%
\usepackage{algorithmicx}%
\usepackage{algpseudocode}%
\usepackage{listings}%
\usepackage{tikz}
\usetikzlibrary{arrows.meta, positioning, shapes.geometric}

\theoremstyle{thmstyleone}%
\theoremstyle{thmstyletwo}%

\theoremstyle{thmstylethree}%

\begin{document}

\textcolor{red}{\textbf{This manuscript has been accepted for publication as a chapter in the book Scientific Machine Learning for Predictive Modeling: Bridging Data-Driven and Physics-Based Approaches in Computational Science and Engineering*, edited by A. Cunha Jr, F. P. Santos, F. A. Rochinha, A. L. G. A. Coutinho, to be published by Springer Nature. The final authenticated version will be available through Springer Nature.}}

\title[Domain-Shift Aware Neural Networks for Unbalance Characterization in Rotating Systems]{Domain-Shift Aware Neural Networks for Unbalance Characterization in Rotating Systems}


\author*[1]{\fnm{Bernardo} \sur{Feijó Junqueira}}\email{bernardo.junqueira@eng.uerj.br}

\author[2]{\fnm{Claudio} \sur{Kiyoshi Umezu}}\email{umezu@unicamp.br}

\author[3]{\fnm{Bruno} \sur{Bilhar Karaziack}}\email{bilhark@cpqd.com.br}

\author[4]{\fnm{Tomaz} \sur{Junior}}\email{tomaz.junior@mail.usf.edu.br}
\author[5]{\fnm{Daniel} \sur{Alves Castello}}\email{castello@mecanica.coppe.ufrj.br}

\affil*[1]{\orgdiv{Department of Mechanical Engineering}, \orgname{Rio de Janeiro State University (UERJ)}, \orgaddress{\street{Fonseca Teles Street, 121}, \city{Rio de Janeiro}, \postcode{20940-200}, \state{RJ}, \country{Brazil}}}

\affil[2]{\orgdiv{Faculty of Agricultural Engineering}, \orgname{University of Campinas (UNICAMP)}, \orgaddress{\street{Av. Cândido Rondon, 501}, \city{Campinas}, \postcode{13083-875}, \state{SP}, \country{Brazil}}}

\affil[3]{\orgdiv{Directorate of Technology and Innovation}, \orgname{CPQD}, \orgaddress{\street{R. Dr. Ricardo Benetton Martins, 1000 - Parque II}, \city{Campinas}, \postcode{13086-902}, \state{SP}, \country{Brazil}}}

\affil[4]{\orgdiv{Department of Computer Engineering}, \orgname{São Francisco University (USF)}, \orgaddress{\street{R. Waldemar César da Silveira, 105 – Jardim Cura D'ars}, \city{Campinas}, \postcode{13045-510}, \state{SP}, \country{Brazil}}}

\affil[5]{\orgdiv{Department of Mechanical Engineering}, \orgname{Federal University of Rio de Janeiro (UFRJ)}, \orgaddress{\city{Rio de Janeiro}, \postcode{21941-594}, \state{RJ}, \country{Brazil}}}


\abstract{This work investigates the application of a domain-shift aware neural network for regression tasks aimed at estimating unbalance masses in rotating shafts under varying operating conditions. Experimental data were collected from a test rig in which a primary shaft, equipped with a flange carrying unbalanced masses, was driven at different rotational speeds, while a secondary shaft could be optionally activated to introduce domain discrepancy. The unbalance masses were positioned at a fixed radial distance, and the dynamic response of the system was recorded using triaxial accelerometers. The inverse problem of mass estimation is formulated within a domain adaptation framework, where the network is trained with a maximum mean discrepancy strategy to align feature representations across source and target distributions. The results demonstrate the effectiveness of explicitly addressing domain shift in improving prediction accuracy, especially when the system’s physical behavior and sources of domain discrepancy are not fully known and fall outside the training conditions. These findings highlight the potential of domain-shift aware models for regression tasks in Structural Health Monitoring.}




\maketitle

\section{Introduction}\label{sec1}

Structural Health Monitoring (SHM) can be understood as the process of implementing damage-assessment strategies for engineering infrastructures \cite{farrar2007}. Over the past decades, the evolution of SHM has closely followed advances in data science. Early SHM approaches were mainly model-based and signal-driven. Around the early 2000s, these methods gradually gave way to data-driven methodologies. More recently, Machine Learning (ML) techniques have developed rapidly, particularly since the 2020s. This progress has enabled SHM systems to overcome long-standing challenges \cite{farrar2025}. These include reduced computational cost and extended inspection capabilities \cite{junqueira2024}. Nevertheless, this progress has also exposed fundamental data-related limitations that continue to constrain further advances in ML-based SHM \cite{farrar2025}.

Among these limitations, one of the most significant challenges in ML-based SHM remains the availability and quality of data. In particular, achieving more accurate damage diagnostics typically requires supervised learning models, which depend on datasets that comprehensively represent all structural states of interest and are accurately labeled. However, in practical applications, such datasets are rarely available \cite{farrar2025,bao2021,malekloo2022}. Moreover, another fundamental challenge is the presence of domain shift, where the data distribution encountered during model training differs from that observed under real operational conditions. Ideally, this would require the structure of interest to be an identical replica of another structure for which labeled damage-state data are available \cite{farrar2025,wang2025,wei2023}. This discrepancy limits the model’s ability to learn representative damage-related patterns. As a result, generalization to unseen conditions is often poor, leading to unreliable diagnostic performance. In safety-critical applications, this may ultimately undermine decision-making \cite{ng2023,qian2024}. In the worst-case scenario, the system operates under baseline conditions that differ significantly from those represented in the training data, while only unlabeled measurements are available, further exacerbating the limitations of supervised ML-based SHM approaches.

To better frame this problem, it is useful to distinguish between two SHM scenarios. The first corresponds to a data-rich setting, referred to here as the source structure, for which labeled measurements under different structural states are available. The second corresponds to the target structure, which represents the system of interest in practical deployments and is typically characterized by limited or unlabeled data, as well as uncertainties in its operating conditions \cite{farrar2025}, leading to domain shift \cite{an2021deep}. The central question then becomes whether knowledge learned from a suitable source structure can be effectively transferred to a target structure while preserving diagnostic reliability. Addressing this challenge is nontrivial, as it requires identifying meaningful similarities between structures operating under different conditions. A common strategy is to learn abstract representations. In this approach, structural responses from different domains are embedded into a shared metric space. In this space, proximity reflects structural similarity \cite{pan2009,li2020,yano2023}. Within this context, domain adaptation has emerged as a widely adopted transfer learning approach in SHM. Its goal is to align source and target feature distributions in a shared latent space. This alignment allows models trained on source data to generalize to target conditions \cite{qian2024,zhu2020,qian2023,schwendemann2021,zhou2021,ghorvei2023,sicilia2023,liu2023,du2021,wang2021}.

Recent studies have demonstrated the effectiveness of domain adaptation techniques in mitigating domain shift in data-driven fault diagnosis of rotating machinery. Efforts have predominantly focused on classification tasks, particularly bearing fault diagnosis, employing discrepancy-based and adversarial strategies. Schwendemann \textit{et al.} \cite{schwendemann2021} introduced a layered maximum mean discrepancy (LMMD) approach combined with an intermediate domain, achieving improved classification accuracy even with limited target-domain data. Similarly, Ghorvei \textit{et al.} \cite{ghorvei2023} employed a convolutional neural network (CNN) with domain adversarial learning and LMMD, reporting promising results in cross-domain bearing fault classification. More advanced frameworks have extended this paradigm to multi-source scenarios, as demonstrated by Liu \textit{et al.} \cite{liu2023}, who proposed a multi-source domain adaptation network integrating discrepancy matching, self-attention mechanisms, and adversarial training to enhance generalization under scarce labeled data. Related studies have also investigated the comparative effectiveness of different discrepancy measures, with An \textit{et al.} \cite{an2021} showing that maximum mean discrepancy (MMD) outperforms divergence-based alternatives such as Kullback–Leibler (KL) and Jensen–Shannon (JS) divergences in rotating machinery fault diagnosis under unlabeled and imbalanced data. More recently, Qian \textit{et al.} \cite{qian2024} proposed an alternative discrepancy metric, termed Variance Discrepancy Representation (VDR), as a replacement for MMD, highlighting ongoing efforts to improve latent-space alignment strategies for bearing fault diagnostics. Despite these advances, the application of domain adaptation to regression tasks in SHM remains comparatively underexplored. One of the few exceptions, despite focusing on a different application domain, is the work of Zhou \textit{et al.} \cite{zhou2021}, who investigated aero-engine parameter deviation estimation using deep domain adaptation with multi-kernel MMD and adversarial domain confusion.

In this context, the objective of this work is to investigate the use of a domain-shift aware neural network (NN) for regression tasks in SHM. The focus is on domain adaptation strategies based on MMD. The proposed approach aims to enable unbalance characterization in rotating systems when discrepancies exist between training and deployment conditions. To this end, experimental data from rotating machinery operating under different conditions are considered to represent source and target structures. Further, only unlabeled data are available for the target domain, allowing the assessment of the model’s ability to transfer learned representations across domains affected by unbalance-induced distribution shifts.

The remainder of this work is organized as follows. Section \textit{Domain-Shift Aware Neural Network} presents the theoretical background, definitions, and mathematical formulation of the proposed model. Section \textit{Physical System} describes the rotating system under investigation and the experimental setup adopted in this work. Section \textit{Domain-Shift Aware Unbalance Characterization} presents and discusses the model performance under different unbalance levels and operating conditions, with particular emphasis on the effects of domain shift. Finally, the \textit{Conclusions} section summarizes the main findings and outlines directions for future works.

\section{Domain-Shift Aware Neural Network}\label{sec2}

In this section, a general formulation of domain-shift aware neural networks based on MMD regularization is presented, independently of the specific application considered later in this work.

In ML–based SHM, NNs are commonly employed as surrogate parametric models to infer the structural condition from measured responses. Let $\pi_{\boldsymbol{\theta}}(\cdot)$ denote an NN parameterized by a collection of trainable parameters $\boldsymbol{\theta}$, which maps an input vector $\boldsymbol{\alpha}_i$ to a predicted structural indicator $\hat{\mathbf{y}}_i = \pi_{\boldsymbol{\theta}}(\boldsymbol{\alpha}_i)$ \cite{silva2025,junqueira2026}. Here, $\boldsymbol{\alpha}_i \in \mathbb{R}^{d_\alpha}$ represents a feature vector extracted from sensor measurements (e.g., time- and frequency-domain signals), where $d_\alpha$ denotes the dimensionality of the input space. The corresponding ground-truth structural parameter(s) are denoted by $\mathbf{y}_i \in \mathbb{R}^{d_y}$, where $d_y$ is the output dimensionality. In the particular case considered in this work, $d_y = 1$, as the objective is to estimate a scalar unbalance level.

In a standard data-driven regression setting, model parameters $\boldsymbol{\theta}$ are learned by minimizing a data-fidelity loss. This loss is typically defined as the Mean Squared Error (MSE) between predictions $\hat{y}_i$ and ground-truth labels $y_i$, expressed as

\begin{equation}
\mathcal{L}_{\text{data}} (\boldsymbol{\theta}) = \frac{1}{N_s} \sum_{i=1}^{N_s} \left( y_i - \hat{y}_i(\boldsymbol{\theta}) \right)^2 ,
\end{equation}
\noindent where $y_i$ denotes the true target value and $N_s$ is the number of labeled samples available for training \cite{bishop2023}. While effective under the assumption that training and deployment data follow the same distribution, this formulation can become insufficient when domain shift is present, as is often the case in practical SHM applications.

To address this limitation, discrepancy-based regularization terms can be incorporated into the learning objective to explicitly reduce the distribution mismatch between source and target domains. Among these approaches, MMD has been widely adopted in domain adaptation due to its non-parametric nature and strong theoretical foundations \cite{farahani2021,briol2019}. Originally introduced as a two-sample statistical test, MMD measures the distance between probability distributions by comparing the mean embeddings of samples in a reproducing kernel Hilbert space (RKHS) induced by a characteristic positive-definite kernel \cite{gretton2012,cheng2021}. Given two independent samples $p$ and $p'$ drawn from a distribution $P$, and two independent samples $q$ and $q'$ drawn from a distribution $Q$, the squared MMD is defined as

\begin{equation}
\mathrm{MMD}_\mathcal{K}^2(P,Q)
=
\mathbb{E}[\mathcal{K}(p,p')]
+
\mathbb{E}[\mathcal{K}(q,q')]
-
2\mathbb{E}[\mathcal{K}(p,q)] .
\label{eq:squared_mmd}
\end{equation}

\noindent Here, $\mathcal{K}(\cdot,\cdot)$ denotes a characteristic positive-definite kernel. The expectation terms $\mathbb{E}[ \cdot]$ quantify similarity between samples drawn from the same domain and across different domains \cite{pomponi2021}.

\begin{figure}[h]
    \centering
    \includegraphics[width=\linewidth]{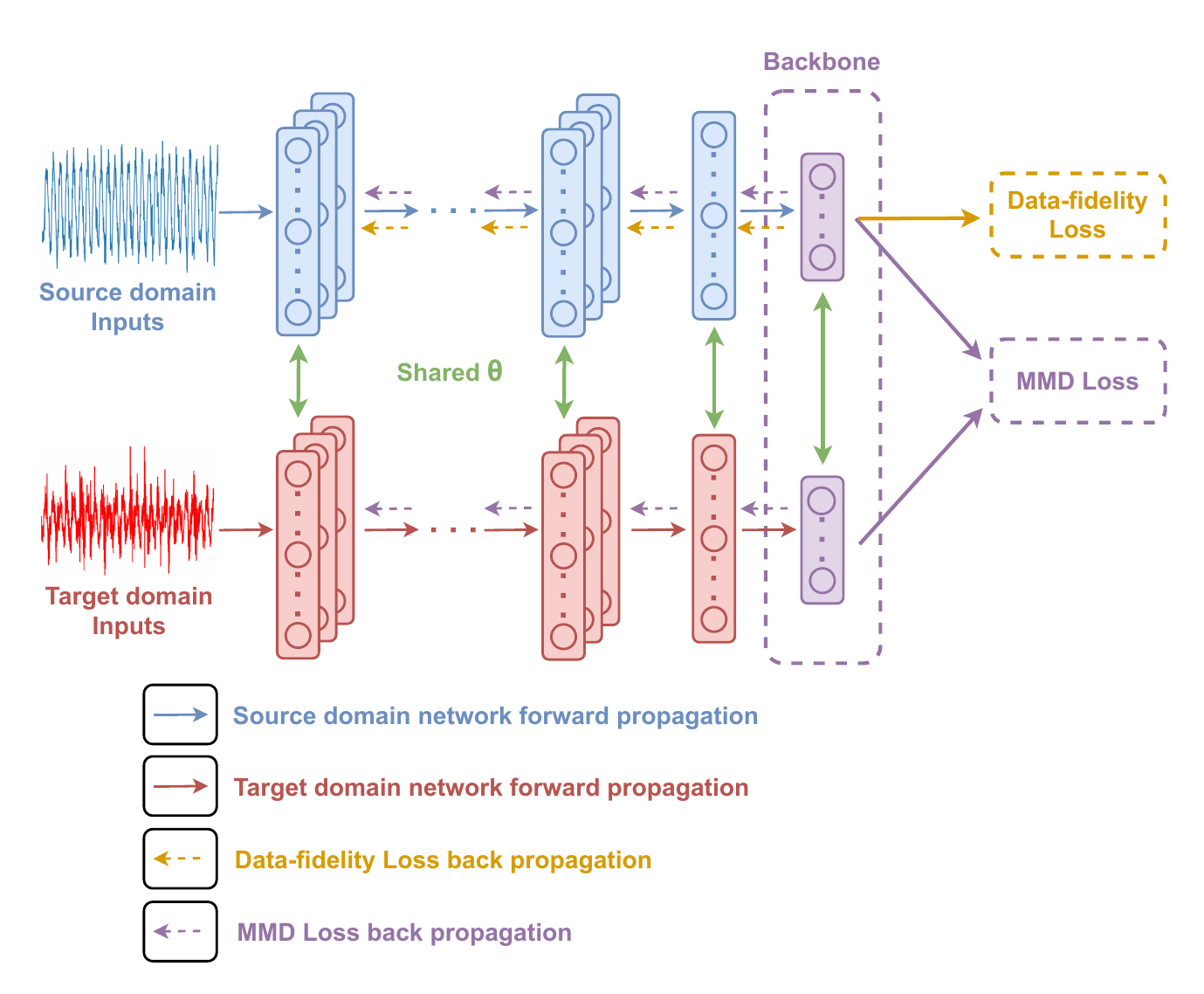}
    \caption{Schematic representation of a domain-shift aware neural network with MMD-based domain adaptation. Source and target domain inputs are processed by two identical network branches sharing parameters $\boldsymbol{\theta}$. The data-fidelity loss is computed only for labeled source-domain samples, while the MMD loss quantifies the discrepancy between source and target latent feature distributions, enabling distribution alignment in the latent space during training.}
    \label{fig:mmd_nn}
\end{figure}

In the context of NNs, this concept can be naturally incorporated into the backbone of the model, i.e., the feature extraction component of the network, typically composed of several hidden layers that map the input measurements into a latent representation space prior to the final prediction layer \cite{bishop2023}. These latent embeddings capture the underlying structural characteristics of the data and provide an appropriate level at which domain discrepancies can be quantified \cite{an2021}. Figure~\ref{fig:mmd_nn} schematically illustrates how discrepancy-based alignment can be performed in the latent space of an NN.

As depicted in Figure~\ref{fig:mmd_nn}, a conventional MMD-based NN architecture consists of two identical branches with shared parameters $\boldsymbol{\theta}$. One branch processes input data from the source domain and is optimized using the data-fidelity loss, while the other processes input data from the target domain. The discrepancy between the source and target feature representations is evaluated in the latent space using the MMD \cite{an2021,qian2023,qian2024}. Based on this formulation, an MMD-based regularization term is introduced to explicitly quantify and penalize the mismatch between source and target feature distributions in the latent space. Let $\{\mathbf{h}_i^s\}_{i=1}^{N_s}$ and $\{\mathbf{h}_j^t\}_{j=1}^{N_t}$ denote the latent feature representations extracted by the shared network from the source and target domains, respectively, where $N_s$ and $N_t$ correspond to the number of available samples in each domain. Here, $\mathbf{h}_i^s$ represents the embedding of the $i$-th source sample, while $\mathbf{h}_j^t$ represents the embedding of the $j$-th target sample in the learned feature space, obtained by the backbone illustrated in Fig.~\ref{fig:mmd_nn}. In practice, the expectation terms in Eq.~\ref{eq:squared_mmd} are not directly accessible and must be approximated from finite samples. This is achieved by replacing the expectations with sample averages over pairs of samples drawn from the source and target domains, resulting in the following empirical expression \cite{qian2023}:

\begin{equation}
\mathcal{L}_{\text{MMD}}(\boldsymbol{\theta}) =
\frac{1}{N_s^2} \sum_{i=1}^{N_s} \sum_{j=1}^{N_s} \mathcal{K}(\mathbf{h}_i^s,\mathbf{h}_j^s)
+
\frac{1}{N_t^2} \sum_{i=1}^{N_t} \sum_{j=1}^{N_t} \mathcal{K}(\mathbf{h}_i^t,\mathbf{h}_j^t)
-
\frac{2}{N_s N_t} \sum_{i=1}^{N_s} \sum_{j=1}^{N_t} \mathcal{K}(\mathbf{h}_i^s,\mathbf{h}_j^t).
\label{eq:loss_mmd}
\end{equation}

The overall training objective is formulated by combining the data-fidelity loss and the MMD regularization term. The resulting domain-shift aware loss $\mathcal{L}(\boldsymbol{\theta})$ is:

\begin{equation}
\mathcal{L}(\boldsymbol{\theta}) =
\mathcal{L}_{\text{data}}(\boldsymbol{\theta})
+
\lambda \mathcal{L}_{\text{MMD}}(\boldsymbol{\theta}),
\label{eq:total_loss}
\end{equation}

\noindent where $\lambda$ is a hyperparameter that controls the trade-off between prediction accuracy on the labeled source domain and distribution alignment between source and target domains. During training, gradients from the data-fidelity loss are backpropagated only through the source-domain branch, while gradients from the MMD loss are backpropagated through both source and target branches due to the shared parameters $\boldsymbol{\theta}$ of the NN (see Fig.~\ref{fig:mmd_nn}). By minimizing the objective function $\mathcal{L}(\boldsymbol{\theta})$ shown in Eq.~\ref{eq:total_loss}, the network is encouraged to learn domain-invariant latent representations, thereby improving generalization performance under domain shift \cite{an2021,qian2023,qian2024}.

\section{Physical System}\label{sec3}

\begin{figure}[h]
    \centering
    \includegraphics[width=0.7\linewidth]{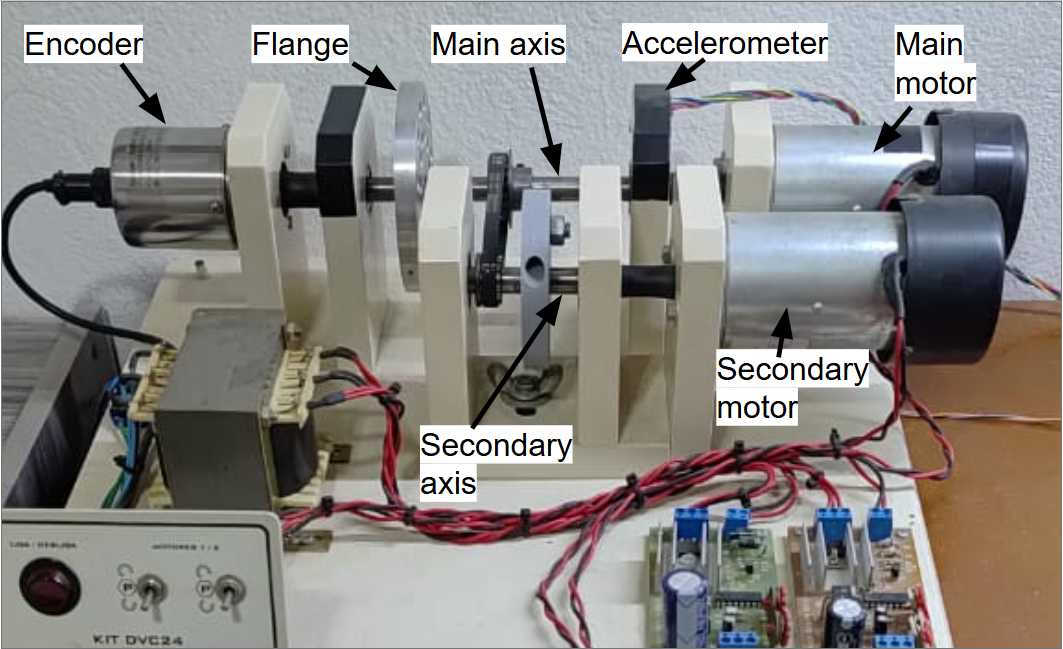}
    \caption{Experimental test rig used to reproduce rotational unbalance phenomena at laboratory scale.}
    \label{fig:physical_system}
\end{figure}

The physical system under investigation is described next. It was specifically designed to reproduce, at a laboratory scale, the main dynamic phenomena associated with unbalance in rotating systems, while preserving fundamental physical relationships observed in real industrial machinery. The experimental test rig used in this work is shown in Fig.~\ref{fig:physical_system}, where the main shaft, the driving electric motor, the support bearings, the flange for mass insertion, and the employed sensors can be identified. The rig consists of a main shaft driven by an electric motor, supported by bearings, and equipped with a flange intended for the controlled insertion of masses. This configuration enables the generation of periodic centrifugal forces proportional to the eccentric mass and to the square of the angular velocity, thereby reproducing the classical physical mechanism of rotational unbalance.

The choice of a triaxial accelerometer allowed the observation of vibrational behavior in orthogonal directions, enabling the capture of both radial components directly associated with unbalance and secondary effects arising from structural coupling and mechanical imperfections. An incremental encoder was used to precisely monitor rotational conditions, ensuring traceability between vibratory features and operational regimens of the system. To better mimic industrial conditions, we added a second shaft driven by an independent motor. The second shaft is coupled to the main shaft by a belt, as shown in Fig.~\ref{fig:physical_system}. The belt connects the left end of the secondary shaft to the midspan of the main shaft. This arrangement introduces additional dynamic interactions, typical of environments where multiple rotating components coexist, resulting in more complex vibration signals with harmonic overlap and increased background noise. Thus, the test rig not only reproduces the isolated phenomenon of unbalance but also includes mechanical interference effects, which are essential for studying the generalization of data-driven models.

\subsection{Experimental Procedure}\label{subsec3-1}

Data acquisition was performed using the previously described triaxial accelerometer (MPU9250), configured to operate within a $\pm$2 g measurement range in order to maximize sensitivity for low-amplitude vibration levels. The sensor was sampled at 1000 Hz (1 kHz), a frequency selected to ensure adequate temporal and spectral resolution for the investigated rotational regimes, considering the expected harmonic content of the signals. 

The highest rotational speed considered in the experiments was 2000 rpm, which corresponds to a fundamental rotational frequency of approximately 33.3 Hz. In rotating machinery, vibration signals typically contain harmonic components at multiples of the rotational frequency, as well as additional contributions from structural resonances and mechanical coupling effects. Considering these factors, relevant vibration components may extend to significantly higher frequencies. Nyquist sampling criterion implies that a 1 kHz sampling rate captures frequencies up to 500 Hz. This range covers the dominant rotational harmonics. It also includes the main structural vibration modes expected in our setup. This choice ensures adequate spectral resolution while maintaining manageable data storage and processing requirements.

Data acquisition was implemented on a Raspberry Pi using C-language routines to ensure temporal stability during high-frequency sampling. Dedicated threads and operating system–level prioritization were employed to minimize timing jitter and sampling interval variability. The acquired data were stored in CSV format, containing the three-axis acceleration components, the instantaneous rotational speed obtained from the incremental encoder, and a timestamp for each sample. Additional details regarding the dataset are provided in Appendix~\ref{secA1}.

The experiments were designed to systematically investigate the relationship between shaft unbalance, rotational speed, and the resulting dynamic complexity of the system. Shaft unbalance was introduced in a controlled manner by modifying the mass distribution on the flange coupled to the main shaft, with the added mass positioned at a radial distance of 40 mm from the shaft center. This radial distance was selected to ensure measurable vibration levels while avoiding excessive mechanical stress. This configuration provides a consistent and physically meaningful increase in centrifugal force across the tested unbalance conditions, improving repeatability and experimental reliability.

\begin{figure}[h]
    \centering
    \includegraphics[width=0.9\linewidth]{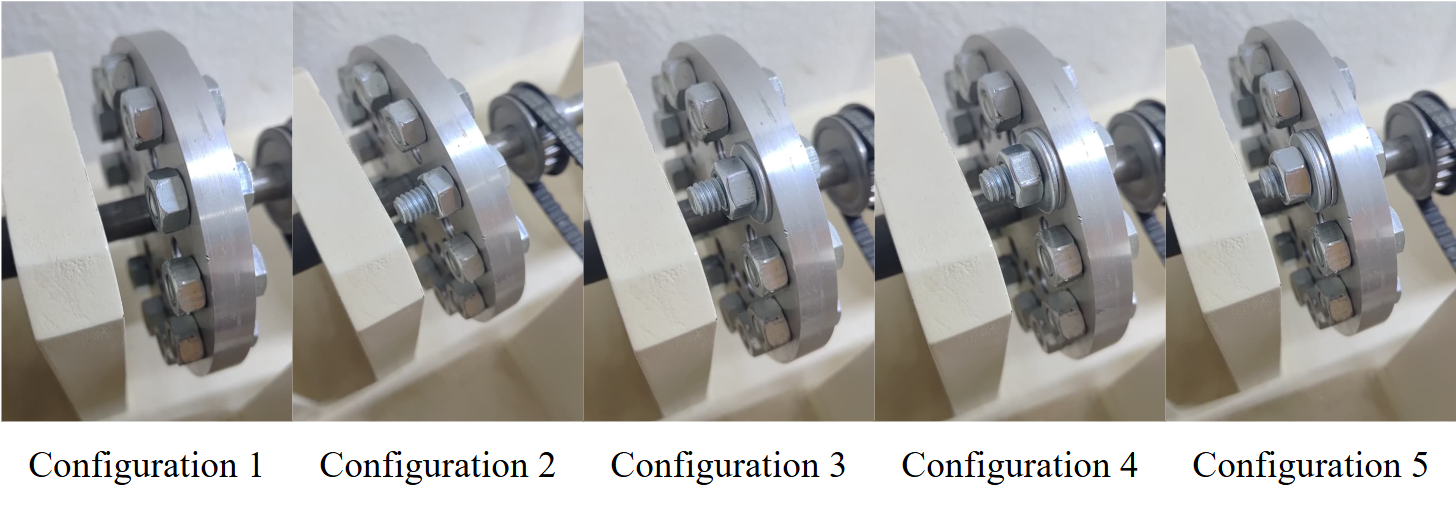}
    \caption{Photos of the unbalanced disk mounted on the first shaft (primary shaft) of the test rig. From configuration 1 to configurations 5 one may observe an increase of added unbalance mass whose values are specified in Table~\ref{tab:unbalance_config}.}
    \label{fig:unbalance_config}
\end{figure}

A set of distinct operating conditions was defined, ranging from a balanced configuration to progressively more severe unbalance states. This experimental design enables the analysis of the continuous evolution of the vibrational response as a function of increasing centrifugal forces, while preserving the physical coherence of the underlying phenomenon. The specific unbalance configurations adopted in the experiments are summarized in Table~\ref{tab:unbalance_config} and schematically illustrated in Fig.~\ref{fig:unbalance_config}.

\begin{table}[h]
\caption{Unbalance configurations adopted in the experimental setup.}
\label{tab:unbalance_config}
\begin{tabular}{@{}cc@{}}
\toprule
Configuration & Added mass (g) \\
\midrule
1 & 0.0 \\
2 & 2.1 \\
3 & 5.0 \\
4 & 7.9 \\
5 & 10.8 \\
\botrule
\end{tabular}
\end{table}

The selected unbalance mass values were defined to generate progressively increasing levels of centrifugal excitation while maintaining safe mechanical operating conditions. The centrifugal force produced by unbalance is proportional to the eccentric mass and to the square of the angular velocity. For this reason, varying mass and rotational speed allows a systematic and controlled escalation of dynamic excitation. Accordingly, we conducted tests at three rotational regimes of the main shaft: 500, 1000, and 2000 rpm, representing low, medium, and high operating conditions. As rotational speed increases, the quadratic dependence of centrifugal force amplifies the effect of each mass increment. This directly influences vibration amplitudes and the corresponding spectral content. This coordinated variation of mass and speed enables a more comprehensive assessment of the progressive amplification of the vibrational response. It also enables assessing the influence of operating conditions on the generalization capability of the machine learning models.

\begin{figure}[h]
    \centering
    \includegraphics[width=0.5\linewidth]{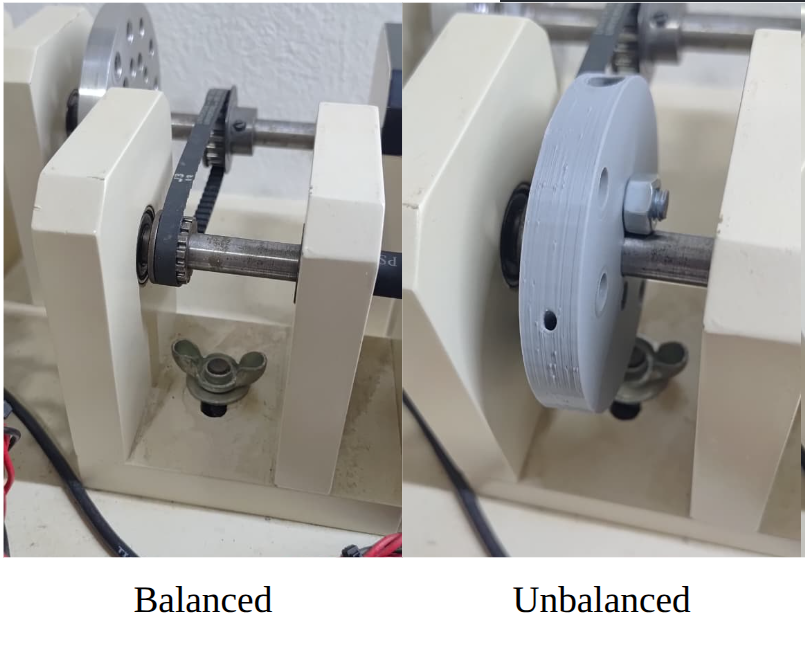}
    \caption{Schematic representation of the second shaft configurations used to introduce domain discrepancy, including balanced and unbalanced conditions.}
    \label{fig:second_shaft}
\end{figure}

Each combination of unbalance level and rotational speed was maintained under steady-state conditions for approximately 10 minutes, ensuring the acquisition of signals were sufficiently long for statistical and spectral analyses, as well as for subsequent segmentation into time windows. This duration was defined to balance operational stability, test repeatability, and sample diversity within the dataset.
Domain discrepancy was introduced by activating a secondary motor. The motor is coupled to the system through a belt transmission. The secondary shaft operates at a fixed rotational ratio with respect to the main shaft and was evaluated under both balanced and unbalanced conditions, as illustrated in Fig.~\ref{fig:second_shaft}. This generates additional dynamic disturbances. It also significantly modifies the system’s vibration signature.

The combination of multiple eccentric mass levels, different rotational regimes, and the presence of dynamic coupling between shafts resulted in a multi-domain experimental dataset, specifically designed to evaluate the robustness and generalization capability of the proposed models. In this way, the experiments were not limited to validation under ideal conditions, but were structured to reproduce, in a controlled manner, typical challenges encountered in the practical application of diagnostic techniques in real rotating systems.

\subsection{Data Processing}\label{subsec3-2}

The raw vibration signals along the three axial directions  were segmented into fixed-length windows of 1000 time points using a sliding-window approach with a step size of 1000, i.e., no overlap between windows, with each segment treated as an independent sample. This choice was made to promote statistical independence between samples, avoiding redundancy that could arise from overlapping segments, at the expense of a reduced number of training samples. 

Prior to feature extraction, the time-domain signals were standardized using $z$-score normalization, with the mean and standard deviation computed exclusively from the source training set. The same standardization parameters were subsequently applied to the source validation set and to all target-domain data, thereby preventing information leakage across domains. This choice is particularly important in domain adaptation settings, as it ensures that no information from the target domain is used during preprocessing, thereby preserving a realistic scenario in which target data are unlabeled and unavailable during training.

To enrich the input representation, frequency-domain information was incorporated by augmenting each window with the magnitude of its Fourier transform. The resulting representation concatenates time-domain and frequency-domain features along the channel dimension, providing complementary spectral information while preserving the original temporal resolution. This representation was adopted as a simple and effective strategy to combine time- and frequency-domain information, allowing the network to jointly exploit temporal patterns and spectral characteristics without requiring more complex feature engineering or model architectures.

Regarding the data-splitting strategy, 80\% of the samples corresponding to unbalance levels of 0 g, 5.0 g, and 10.8 g were allocated to the training set. The remaining 20\% of these samples were combined with all samples associated with unbalance levels of 2.1 g and 7.9 g, which were not used during training. This combined subset was then stratified according to the number of samples per unbalance level and randomly divided into validation and test sets in equal proportions (50/50). Consequently, the validation and test sets share the same distribution of unbalance levels while containing mutually exclusive samples, enabling both hyperparameter tuning and an unbiased assessment of the model’s generalization performance under previously unseen operating conditions.

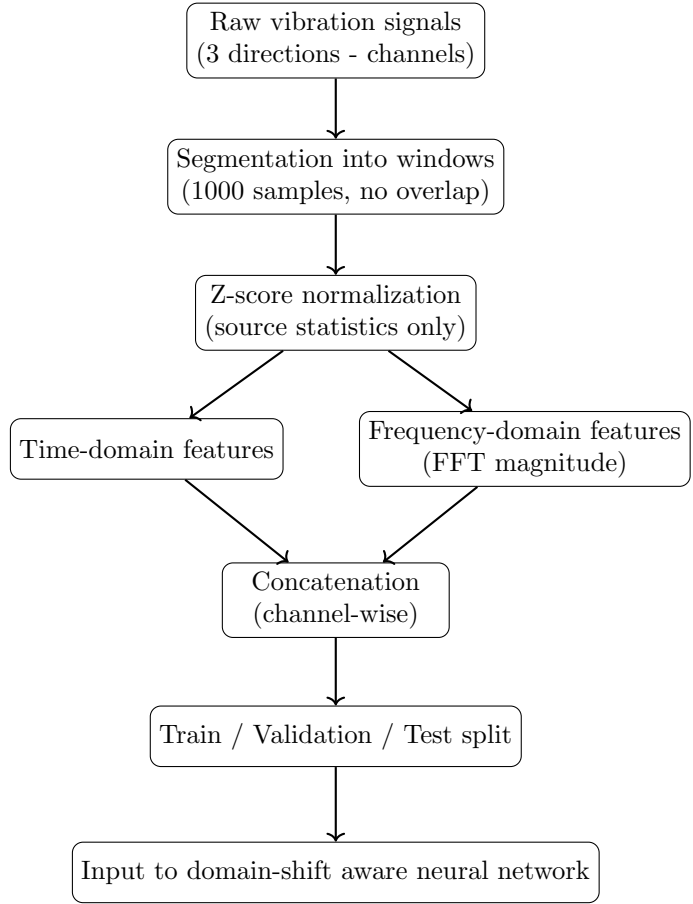
\begin{figure}[h]
\centering
\begin{tikzpicture}[
node distance=1.8cm,
every node/.style={draw, rectangle, rounded corners, align=center, minimum width=3cm, minimum height=0.8cm},
arrow/.style={->, thick}
]

\node (raw) {Raw vibration signals\\(3 directions - channels)};
\node (window) [below of=raw] {Segmentation into windows\\(1000 samples, no overlap)};
\node (norm) [below of=window] {Z-score normalization\\(source statistics only)};
\node (time) [below of=norm, xshift=-2.5cm] {Time-domain features};
\node (freq) [below of=norm, xshift=2.5cm] {Frequency-domain features\\(FFT magnitude)};
\node (concat) [below of=norm, yshift=-2cm] {Concatenation\\(channel-wise)};
\node (split) [below of=concat] {Train / Validation / Test split};
\node (model) [below of=split] {Input to domain-shift aware neural network};

\draw[arrow] (raw) -- (window);
\draw[arrow] (window) -- (norm);
\draw[arrow] (norm) -- (time);
\draw[arrow] (norm) -- (freq);
\draw[arrow] (time) -- (concat);
\draw[arrow] (freq) -- (concat);
\draw[arrow] (concat) -- (split);
\draw[arrow] (split) -- (model);

\end{tikzpicture}
\caption{Overview of the data processing pipeline adopted for the proposed MMD-based domain adaptation model for unbalance estimation.}
\label{fig:data_pipeline}
\end{figure}

The same pre-processing and splitting pipeline was consistently applied across all experiments and methods under investigation, ensuring that observed performance differences arise solely from the learning strategies rather than data handling artifacts. A schematic overview of the data processing pipeline is provided in Fig.~\ref{fig:data_pipeline}.

\section{Results: Domain-Shift Aware Unbalance Characterization}\label{sec4}

This section provides a comprehensive evaluation of the proposed domain-shift aware framework for unbalance characterization. The results are analyzed in terms of training stability, domain alignment effectiveness, and predictive accuracy on unseen target-domain conditions. Particular attention is given to the interplay between regression and MMD losses, as well as to the robustness of the learned representations across different rotational speed regimes. To facilitate a structured comparison, the results are presented separately for each operating condition.

\subsection{Proposed Model}\label{subsec4-0}

Building upon the general framework introduced in Section~\ref{sec2}, the approach is here tailored to the unbalance estimation problem considered in this work. The proposed model employs a shared one-dimensional convolutional neural network (CNN) backbone for both source and target domains, as summarized in Table~\ref{tab:architecture}. The backbone consists of three convolutional blocks with increasing filter depths, followed by global average pooling and fully connected layers, producing a compact 64-dimensional feature representation and a scalar regression output. As illustrated in Fig.~\ref{fig:mmd_nn}, this backbone is applied in a twin-network configuration, where both source and target samples are processed by identical, weight-shared networks. While only source-domain samples contribute to the regression loss, the extracted feature representations from both domains are jointly used to compute the MMD loss, encouraging domain-invariant feature learning.

\begin{table}[h]
\caption{Architecture of the CNN backbone used for feature extraction and regression.}
\label{tab:architecture}
\begin{tabular}{@{}llll@{}}
\toprule
Layer & Type & Output Shape & Parameters \\
\midrule
Input & InputLayer & $(1000 \times 6)$ & 0 \\

Conv1 & Conv1D (32 filters, $k=5$) & $(1000 \times 32)$ & 992 \\

Pool1 & MaxPooling1D ($p=2$) & $(500 \times 32)$ & 0 \\

Conv2 & Conv1D (64 filters, $k=3$) & $(500 \times 64)$ & 6,208 \\

Pool2 & MaxPooling1D ($p=2$) & $(250 \times 64)$ & 0 \\

Conv3 & Conv1D (128 filters, $k=3$) & $(250 \times 128)$ & 24,704 \\

GAP & GlobalAveragePooling1D & $(128)$ & 0 \\

FC1 & Dense (256 units) & $(256)$ & 33,024 \\

FC2 & Dense (128 units) & $(128)$ & 32,896 \\

Feature Layer & Dense (64 units) & $(64)$ & 8,256 \\

Output & Dense (1 unit) & $(1)$ & 65 \\
\botrule
\end{tabular}
\footnotetext{Note: $k$ denotes the kernel size of the convolutional layers, and $p$ denotes the pooling size of the max-pooling layers. All convolutional layers use ReLU activation and same padding.}
\end{table}

The adopted architecture reflects a trade-off between representational capacity and generalization, following commonly used configurations for vibration-based fault diagnosis under domain adaptation settings \cite{qian2023, qian2024}. In particular, moderate-depth CNN backbones have been shown to be effective in extracting localized temporal patterns while avoiding overfitting in relatively limited datasets.

To assess the sensitivity of the proposed approach to architectural choices, a preliminary hyperparameter tuning was conducted over the network depth and width. The explored configurations ranged from shallower models with two convolutional layers (32 and 64 filters) and two fully connected layers (64 and 1 units), to deeper architectures with up to four convolutional layers (32, 64, 128, and 256 filters) and four fully connected layers (512, 256, 128, 64, and 1 units). Additionally, different feature extraction depths were evaluated for coupling with the MMD-based alignment mechanism. Furthermore, different activation functions, including ReLU and hyperbolic tangent (tanh), were investigated to assess their impact on representation learning and training stability.

The selected architecture (Table~\ref{tab:architecture}) provided the best compromise between predictive performance and training stability across operating conditions. Deeper architectures did not yield consistent improvements and, in some cases, led to increased overfitting and training instability, particularly under stronger domain shift conditions. Similarly, the use of tanh activations resulted in slower convergence and slightly degraded performance compared to ReLU. These observations suggest that the proposed results are not overly sensitive to minor architectural variations, while still benefiting from a sufficiently expressive yet compact model.

The network was trained using the Adam optimizer with a fixed learning rate of $10^{-4}$, selected to ensure stable convergence across all experimental conditions. For the computation of the MMD loss, a Student’s \emph{t}-kernel was adopted as the characteristic positive-definite  kernel $\mathcal{K}$ present in Eqs.~\ref{eq:squared_mmd} and~\ref{eq:loss_mmd}. This choice was motivated by the findings reported in \cite{qian2024}, which demonstrate that the Student kernel provides improved sensitivity to distributional discrepancies in deep feature spaces when compared to commonly used Gaussian kernels, particularly under complex and heavy-tailed data distributions. From a functional perspective, the heavier tails of the Student’s \emph{t}-kernel assign non-negligible similarity values even to samples that are farther apart in the feature space, thereby enhancing the ability of the MMD to capture discrepancies in regions with high variability or sparse support. The Student’s \emph{t}-kernel $\mathcal{K}(\mathbf{h}_i^s, \mathbf{h}_j^t)$ employed in the MMD loss is defined as
\begin{equation}
\mathcal{K}(\mathbf{h}_i^s, \mathbf{h}_j^t)
=
C_{\nu}
\left(\nu + \lVert \mathbf{h}_i^s - \mathbf{h}_j^t \rVert_2^2 \right)^{-\frac{\nu+1}{2}},
\end{equation}
\noindent where $\nu$ is the degrees-of-freedom parameter controlling the tail behavior of the kernel. Specifically, $\nu$ directly controls the heaviness of the kernel tails: smaller values of $\nu$ produce heavier tails, assigning higher similarity to distant samples, whereas larger values lead to a more rapidly decaying kernel, approaching Gaussian-like behavior. Here, $C_{\nu}$ is a normalization constant given by
\begin{equation}
C_{\nu} = \frac{\Gamma\left(\frac{\nu+1}{2}\right)}{\Gamma\left(\frac{\nu}{2}\right)\sqrt{\nu\pi}},
\end{equation}
where $\Gamma(\cdot)$ denotes the Gamma function, a standard extension of the factorial to real-valued arguments. This normalization ensures that the kernel satisfies the properties of a valid positive-definite function.

To balance the contribution of the domain alignment loss during training, a progressive weighting strategy based on a sigmoid-shaped lambda scheduler was employed \cite{ganin2016}. The MMD loss weight $\lambda$, shown in Eq.~\ref{eq:total_loss}, is defined as a function of the normalized training progress $\epsilon$ according to

\begin{equation}
\lambda(\epsilon) =
\lambda_{\max}
\left(
\frac{2}{1 + \exp(-\gamma \epsilon)} - 1
\right),
\end{equation}
\noindent where $\gamma$ and $\lambda_{max}$ were set to 10. The parameter $\gamma$ controls the growth rate of the scheduling curve, with larger values leading to a steeper transition from low to high $\lambda$ values, while smaller values result in a more gradual increase. The parameter $\lambda_{\max}$ defines the upper bound of the domain alignment contribution, directly controlling the relative importance of the MMD loss in the final stages of training. The normalized training progress $\epsilon$ is defined as 

\begin{equation}
\epsilon = \frac{e}{E},
\end{equation}
\noindent where $e$ denotes the current training epoch and $E$ is the total number of training epochs. This formulation allows the influence of the MMD loss to increase gradually throughout training, starting from zero and asymptotically approaching $\lambda_{max}$. In the early training stages, the optimization is dominated by the regression objective, facilitating stable feature learning, while stronger domain alignment is enforced in later stages. In other words, the scheduling strategy mitigates potential instability that could arise from enforcing strong domain alignment at early stages, when feature representations are still poorly structured.

The experimental results are presented separately for different rotational speed conditions in order to analyze the robustness of the proposed domain adaptation framework under varying operating regimes.

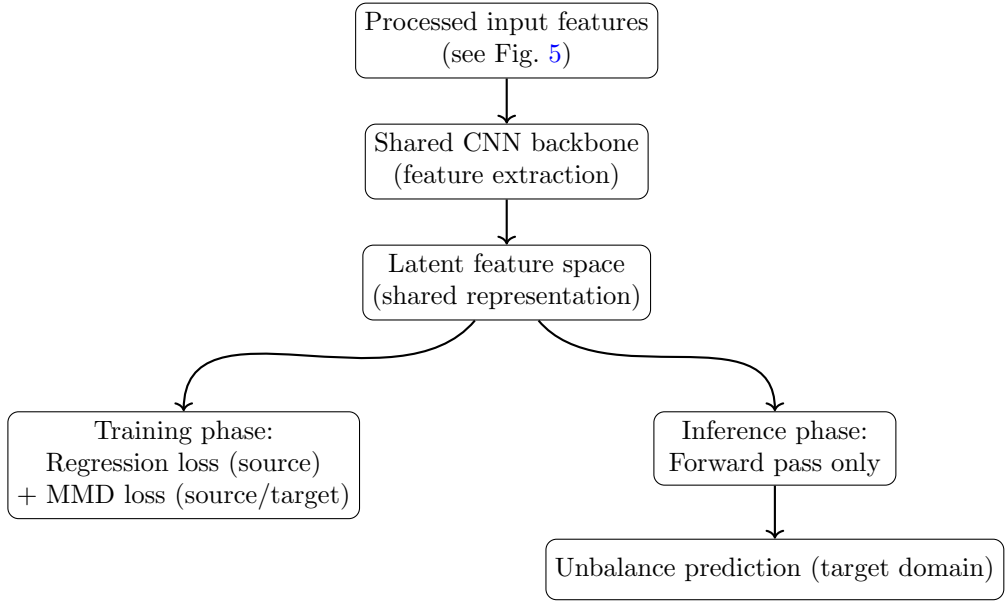
\begin{figure}[h]
\centering
\begin{tikzpicture}[
node distance=1.6cm,
every node/.style={draw, rectangle, rounded corners, align=center, minimum width=3.2cm, minimum height=0.8cm},
arrow/.style={->, thick}
]

\node (prep) {Processed input features\\(see Fig.~\ref{fig:data_pipeline})};
\node (cnn) [below of=prep] {Shared CNN backbone\\(feature extraction)};
\node (feat) [below of=cnn] {Latent feature space\\(shared representation)};

\node (train) [below left=1.2cm and 0.03cm of feat] {Training phase:\\Regression loss (source)\\+ MMD loss (source/target)};
\node (infer) [below right=1.2cm and 0.03cm of feat] {Inference phase:\\Forward pass only};

\node (out) [below of=infer] {Unbalance prediction (target domain)};

\draw[arrow] (prep) -- (cnn);
\draw[arrow] (cnn) -- (feat);
\draw[arrow] (feat) to[out=230, in=90] (train);
\draw[arrow] (feat) to[out=310, in=90] (infer);
\draw[arrow] (infer) -- (out);

\end{tikzpicture}
\caption{End-to-end workflow of the proposed domain-shift aware methodology for unbalance estimation. The diagram distinguishes between the training phase, where regression and MMD losses are jointly optimized, and the inference phase, where the trained model is used to predict unbalance levels in the target domain.}
\label{fig:full_pipeline}
\end{figure}

From a workflow perspective, the proposed methodology can be interpreted as an end-to-end pipeline that integrates data preprocessing, feature extraction, and domain-adaptive learning. The initial stages of signal processing and feature construction follow the procedure illustrated in Fig.~\ref{fig:data_pipeline}. The resulting inputs are then processed by the domain-shift aware neural network, whose general architecture and MMD-based alignment mechanism are depicted in Fig.~\ref{fig:mmd_nn}, while its specific instantiation for the unbalance estimation problem is detailed in Table~\ref{tab:architecture}. Within this framework, the shared CNN backbone maps the inputs into a latent feature space, where domain alignment is enforced through the MMD loss while regression is guided by labeled source-domain data. The complete workflow, from processed measurements to final prediction, is summarized in Fig.~\ref{fig:full_pipeline}.

\subsection{Results across operating regimes}

This section evaluates the proposed domain-shift aware model across three rotational speeds (500, 1000, and 2000 RPM) and two structural configurations (balanced and unbalanced second shaft). Rather than analyzing each case in isolation, the results are presented with emphasis on how model behavior evolves with increasing rotational speed and structural complexity.

To provide a consistent basis for comparison, the analysis is organized around three complementary aspects:

\begin{itemize}
    \item[(i)] \textit{Training dynamics}: The learning process is examined through the evolution of the loss function, including both training and validation curves. The individual contributions of the data and the MMD losses are reported, allowing the assessment of prediction accuracy and domain alignment throughout training, as well as their relative influence on convergence behavior.

    \item[(ii)] \textit{Predictive accuracy in the target domain}: Model performance is evaluated using predicted versus true unbalance values in the target domain, including the mean prediction for each unbalance level. Quantitative metrics, such as root mean square error (RMSE), mean absolute error (MAE), and coefficient of determination ($R^2$), are reported to summarize regression accuracy. For each operating condition, results are presented both with and without the MMD-based domain alignment strategy (i.e., $\lambda = 0$ in Eq.~\ref{eq:total_loss}), enabling a direct comparison of cross-domain generalization performance.

    \item[(iii)] \textit{Structure of the learned feature space}: The internal representation learned by the model is analyzed using t-distributed Stochastic Neighbor Embedding (t-SNE) \cite{cai2022}, a nonlinear dimensionality reduction technique that projects high-dimensional features into a two-dimensional space while preserving local neighborhood relationships. These visualizations provide a qualitative view of how samples are organized with respect to unbalance levels and how source and target domains are positioned relative to each other in the latent space.
\end{itemize}

The following subsections present the results for each operating regime using this common analysis framework.

\subsubsection{Low-speed regime: 500 RPM}

At 500 RPM, the proposed model exhibits limited generalization capability across domains. This operating regime is characterized by increased sensitivity to both domain shift and structural variability, making it a challenging scenario for cross-domain prediction.

\begin{figure}[H]
    \centering
    \includegraphics[width=\linewidth]{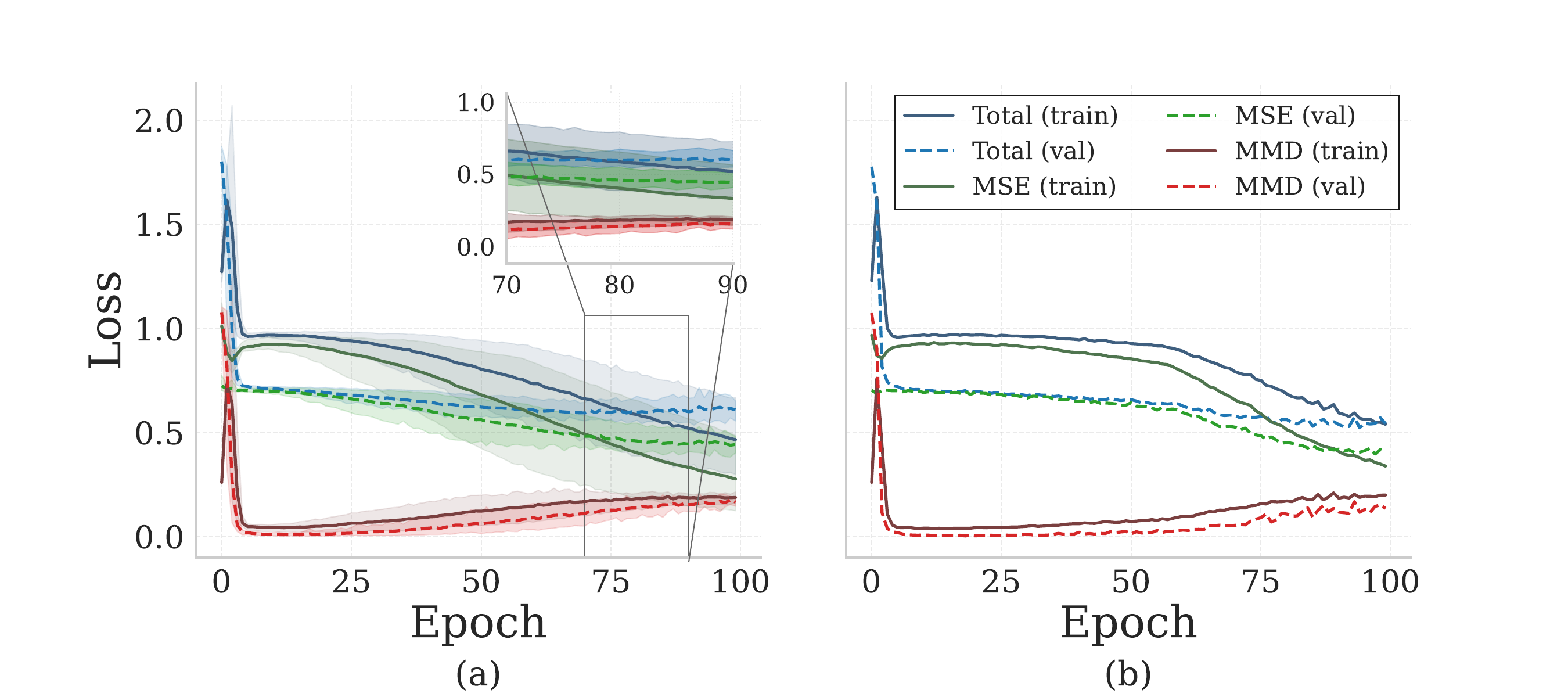}
    \caption{Training and validation loss evolution for the domain-shift aware model trained at the 500 RPM regime with a balanced second-shaft condition. (a) Mean loss curves across 10 independent training runs, with shaded regions indicating the 5th–95th percentile range; the inset zooms into the late-training regime. (b) Loss curves of the best-performing run, defined as the run with the lowest final validation loss. Solid and dashed lines correspond to training and validation losses, respectively.}
    \label{fig:loss_500_balanced}
\end{figure}

\begin{figure}[H]
    \centering
    \includegraphics[width=\linewidth]{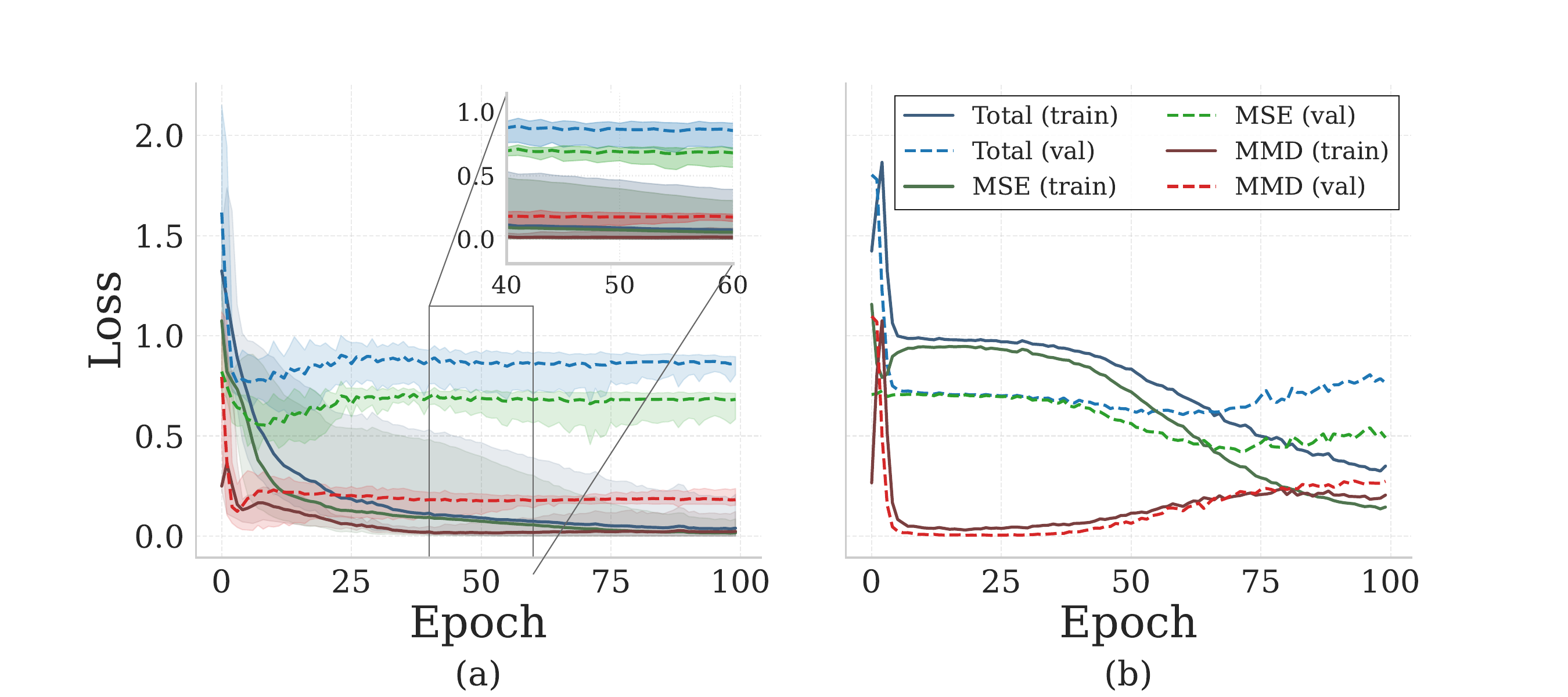}
    \caption{Training and validation loss evolution for the domain-shift aware model trained at the 500 RPM regime with an unbalanced second-shaft condition. (a) Mean loss curves across 10 independent training runs, with shaded regions indicating the 5th–95th percentile range; the inset zooms into the \textcolor{red}late-training regime. (b) Loss curves of the best-performing run, defined as the run with the lowest final validation loss. Solid and dashed lines correspond to training and validation losses, respectively.}
    \label{fig:loss_500_unbalanced}
\end{figure}

\paragraph{\textbf{(i) Training dynamics}}
The loss evolution in Fig.~\ref{fig:loss_500_balanced} indicates that the MMD component remains consistently low, suggesting effective domain alignment. The optimization process is dominated by the data loss, reflecting limited predictive capability on the target domain. This behavior becomes more pronounced under the unbalanced second-shaft condition (Fig.~\ref{fig:loss_500_unbalanced}), where a persistent gap between training and validation curves indicates reduced generalization, consistent with stronger overfitting and increased sensitivity to structural discrepancy.

\begin{figure}[H]
    \centering
    \includegraphics[width=0.5\linewidth]{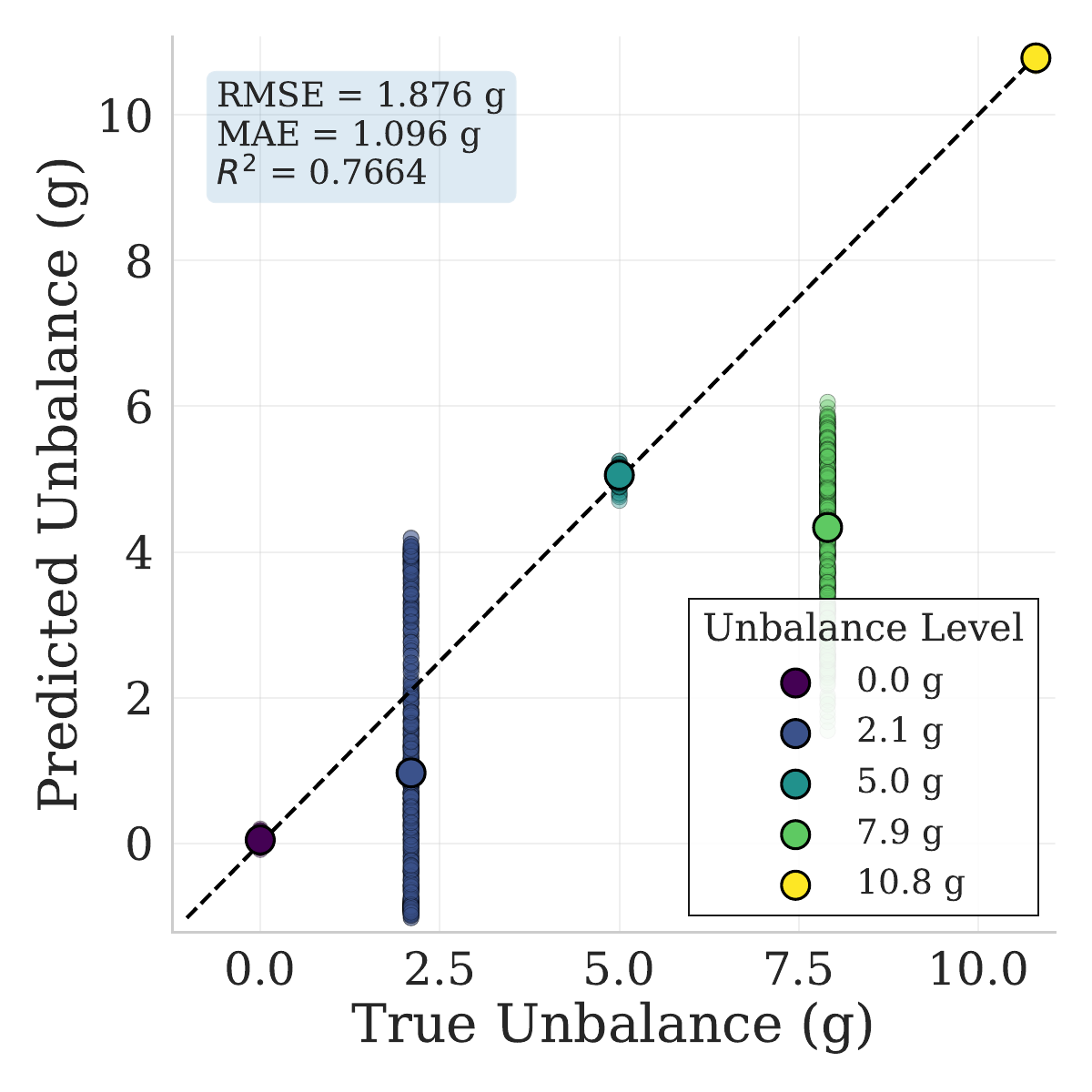}
    \caption{Predicted versus true unbalance values for the target domain at 500 RPM under the balanced second-shaft condition using the domain-shift aware model. Each color represents a distinct unbalance level. Translucent markers indicate individual predictions, while larger markers denote the mean predicted value for each level. The dashed line represents ideal prediction. The dispersion of predictions around the ideal line reflects the estimation error for each unbalance level, with larger spread indicating lower accuracy and higher uncertainty. Systematic deviations of the mean predictions from the ideal line reveal bias in the model response. These effects directly impact the reported performance metrics, where increased dispersion and bias lead to higher RMSE and MAE values and a reduction in $R^2$. Consistent clustering along the ideal line indicates reliable generalization.
    }
    \label{fig:predictions_500}
\end{figure}

\begin{figure}[H]
    \centering
    \includegraphics[width=0.5\linewidth]{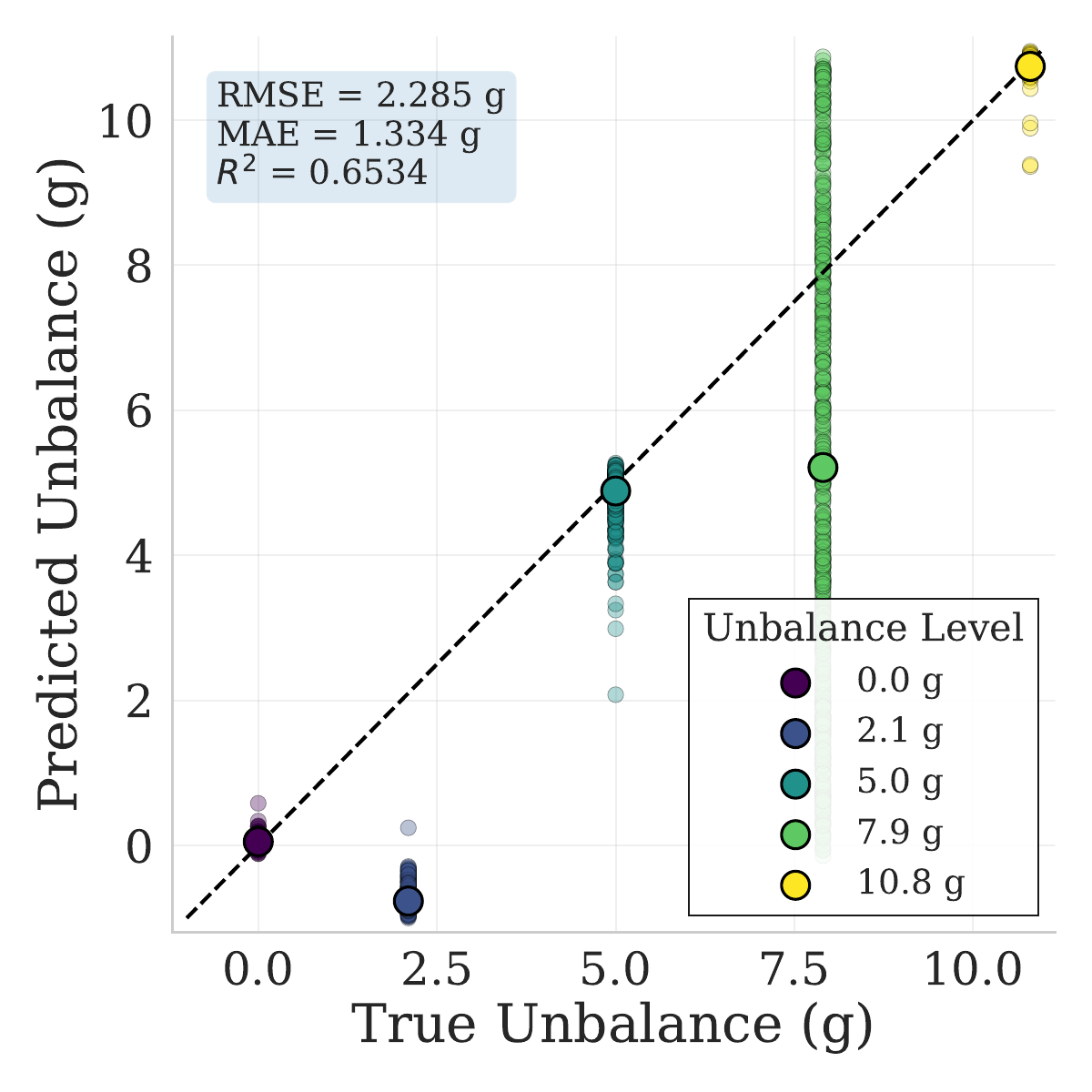}
    \caption{Predicted versus true unbalance values for the target domain at 500 RPM under the unbalanced second-shaft condition using the domain-shift aware model. Each color represents a distinct unbalance level. Translucent markers indicate individual predictions, while larger markers denote the mean predicted value for each level. The dashed line represents ideal prediction. The dispersion of predictions around the ideal line reflects the estimation error for each unbalance level, with larger spread indicating lower accuracy and higher uncertainty. Systematic deviations of the mean predictions from the ideal line reveal bias in the model response. These effects directly impact the reported performance metrics, where increased dispersion and bias lead to higher RMSE and MAE values and a reduction in $R^2$. Consistent clustering along the ideal line indicates reliable generalization.
    }
    \label{fig:predictions_500_unb}
\end{figure}

\paragraph{\textbf{(ii) Predictive performance}}
These limitations are directly reflected in the predictive results. Under the balanced second-shaft condition (Fig.~\ref{fig:predictions_500}), the model exhibits noticeable dispersion, particularly for unbalance levels not seen during training, leading to only moderate agreement with the ground truth (RMSE = 1.876 g, MAE = 1.096 g, $R^2$ = 0.7664). This behavior further deteriorates in the unbalanced case (Fig.~\ref{fig:predictions_500_unb}), where dispersion increases across all levels, highlighting reduced robustness to domain shift under increased structural complexity.

\begin{figure}[H]
    \centering
    \includegraphics[width=0.6\linewidth]{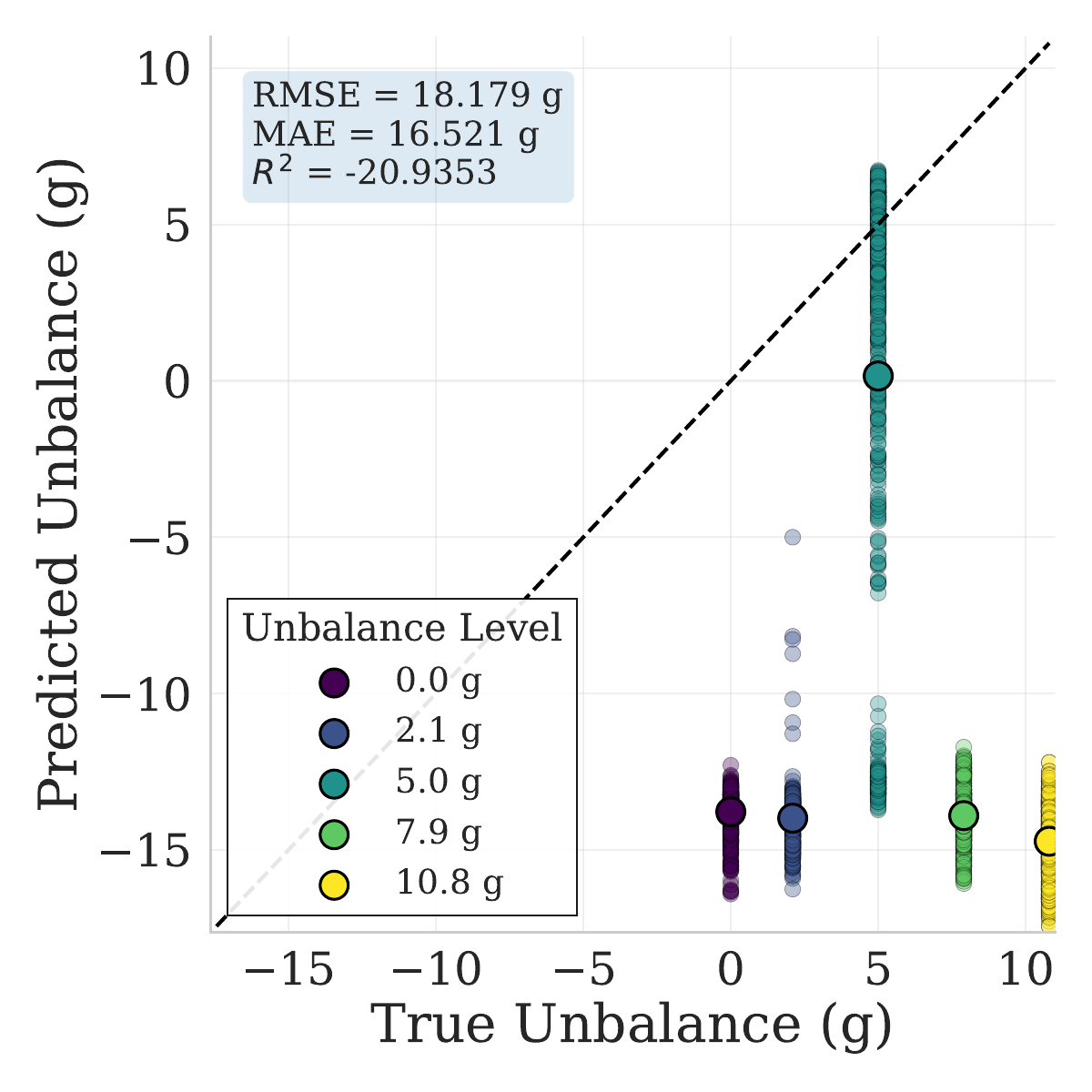}
    \caption{Predicted versus true unbalance values at 500 RPM under the balanced second-shaft condition for the baseline model without domain alignment strategy (i.e., $\lambda = 0$ in Eq.~\ref{eq:total_loss}). Each color represents a distinct unbalance level. Translucent markers indicate individual predictions, while larger markers denote the mean predicted value for each level. The dashed line represents ideal prediction. The dispersion of predictions around the ideal line reflects the estimation error for each unbalance level, with larger spread indicating lower accuracy and higher uncertainty. Systematic deviations of the mean predictions from the ideal line reveal bias in the model response. These effects directly impact the reported performance metrics, where increased dispersion and bias lead to higher RMSE and MAE values and a reduction in $R^2$. Consistent clustering along the ideal line indicates reliable generalization.
    }
    \label{fig:predictions_500_0}
\end{figure}

\begin{figure}[H]
    \centering
    \includegraphics[width=0.6\linewidth]{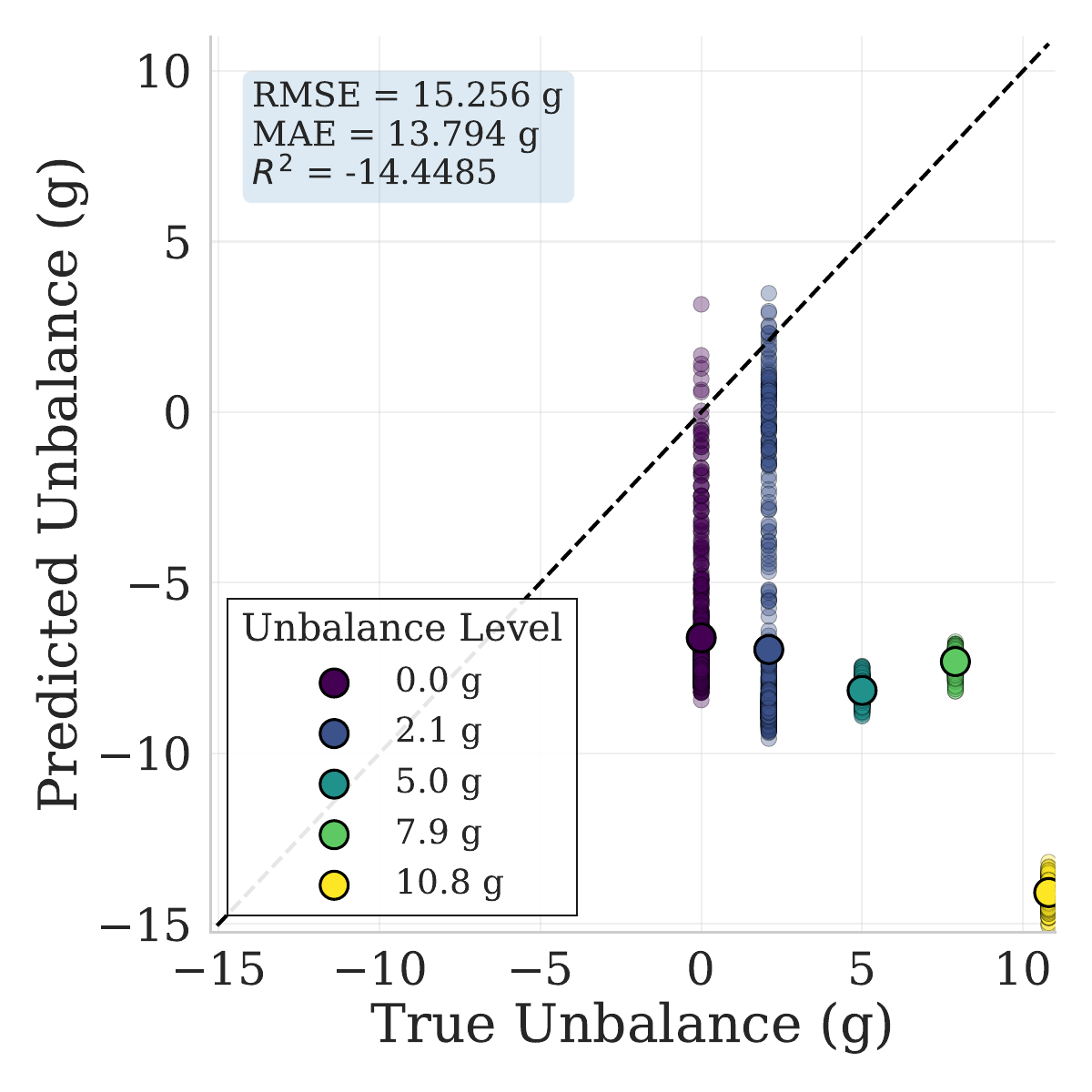}
    \caption{Predicted versus true unbalance values at 500 RPM under the unbalanced second-shaft condition for the baseline model without domain alignment strategy (i.e., $\lambda = 0$ in Eq.~\ref{eq:total_loss}). Each color represents a distinct unbalance level. Translucent markers indicate individual predictions, while larger markers denote the mean predicted value for each level. The dashed line represents ideal prediction. The dispersion of predictions around the ideal line reflects the estimation error for each unbalance level, with larger spread indicating lower accuracy and higher uncertainty. Systematic deviations of the mean predictions from the ideal line reveal bias in the model response. These effects directly impact the reported performance metrics, where increased dispersion and bias lead to higher RMSE and MAE values and a reduction in $R^2$. Consistent clustering along the ideal line indicates reliable generalization.
    }
    \label{fig:predictions_500_unb_0}
\end{figure}

The comparison with the baseline model highlights the importance of domain alignment. Figures~\ref{fig:predictions_500_0} and \ref{fig:predictions_500_unb_0} shows that without the MMD loss, predictions become highly unstable, with large dispersion and physically inconsistent values, leading to a drastic increase in error (e.g., RMSE = 18.179 g in the balanced case).

\begin{figure}[H]
    \centering
    \includegraphics[width=\linewidth]{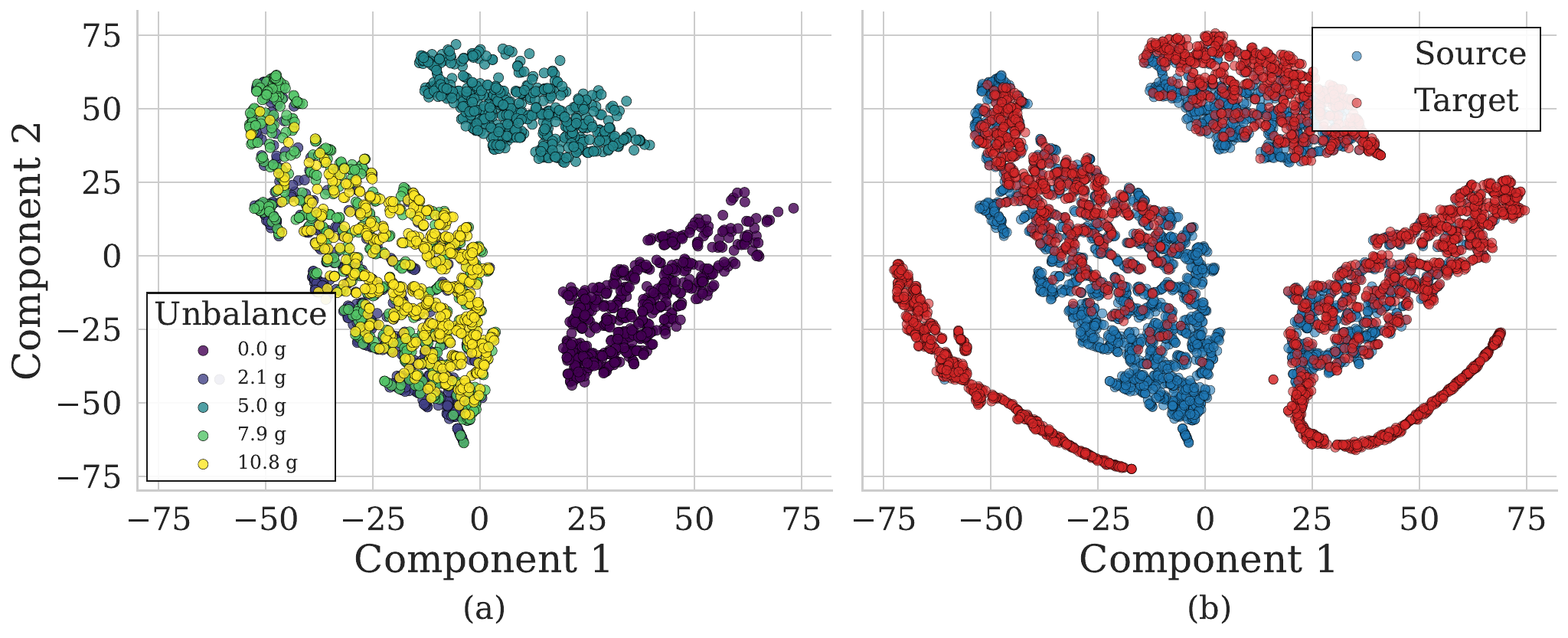}
    \caption{Two-dimensional t-SNE projection of the backbone feature representations extracted by the model at 500 RPM under the balanced second-shaft condition. (a) Source-domain features, where each color represents a cluster associated with a specific unbalance level. Well-separated and compact clusters indicate high feature discriminability and facilitate reliable regression, including interpolation for unseen levels. In contrast, significant overlap between clusters reflects reduced separability, which can impair the model’s ability to distinguish between operating conditions and degrade predictive performance. (b) Embedded feature distributions of source and target domains, highlighting their spatial relationship in the learned feature space. A higher degree of overlap between domains suggests more effective domain alignment, provided that feature separability is preserved for accurate cross-domain regression.
    }
    \label{fig:tsne_500}
\end{figure}

\paragraph{\textbf{(iii) Feature space structure}}
The t-SNE projections in Fig.~\ref{fig:tsne_500} provide insight into the origin of these limitations. In the source-domain embedding (Fig.~\ref{fig:tsne_500}(a)), the feature space exhibits limited separation among different unbalance levels, indicating weak discriminative structure. In the joint embedding (Fig.~\ref{fig:tsne_500}(b)), source and target domains largely overlap, confirming that domain alignment reduces distributional discrepancy. However, clusters associated with different unbalance levels remain strongly intermixed, preventing the formation of a feature space suitable for reliable regression. This lack of structure is consistent with the poor generalization observed in Fig.~\ref{fig:predictions_500}.

\begin{figure}[H]
    \centering
    \includegraphics[width=\linewidth]{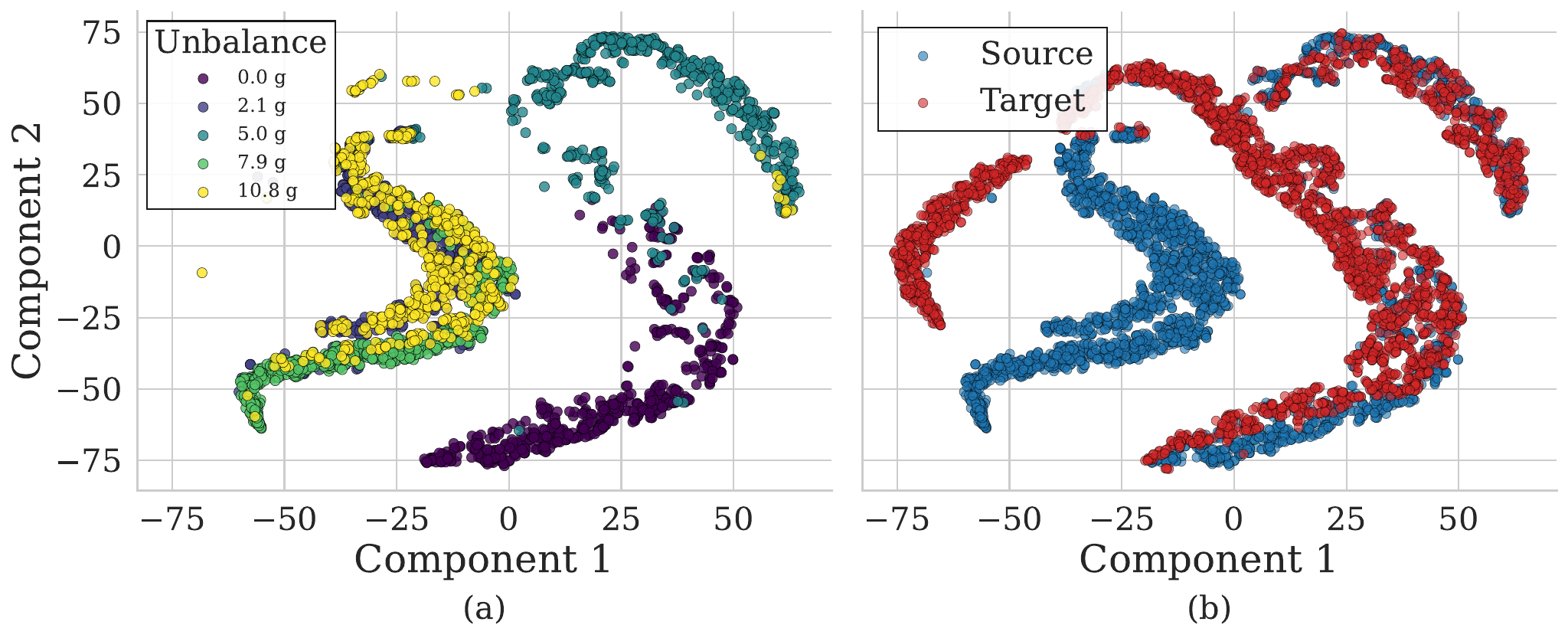}
    \caption{Two-dimensional t-SNE projection of the backbone feature representations extracted by the model at 500 RPM under the unbalanced second-shaft condition. (a) Source-domain features, where each color represents a cluster associated with a specific unbalance level. Well-separated and compact clusters indicate high feature discriminability and facilitate reliable regression, including interpolation for unseen levels. In contrast, significant overlap between clusters reflects reduced separability, which can impair the model’s ability to distinguish between operating conditions and degrade predictive performance. (b) Embedded feature distributions of source and target domains, highlighting their spatial relationship in the learned feature space. A higher degree of overlap between domains suggests more effective domain alignment, provided that feature separability is preserved for accurate cross-domain regression.
    }
    \label{fig:tsne_500_unb}
\end{figure}

This effect is intensified under the unbalanced condition (Fig.~\ref{fig:tsne_500_unb}), where feature distributions become even more entangled. The absence of well-defined clusters explains the observed prediction errors.

Overall, the results at 500 RPM indicate that poor generalization arises from the combined effect of weak feature separability and limited domain alignment, particularly in the presence of structural asymmetry. This reflects intrinsic limitations of the data in this low-speed regime, which hinder the formation of a discriminative and transferable feature representation.

\subsubsection{Medium-speed regime: 1000 RPM}

At 1000 RPM, the proposed model exhibits strong generalization capability across domains. In contrast to the low-speed regime, this operating condition provides more informative patterns, enabling both effective feature learning and robust domain alignment.

\begin{figure}[H]
    \centering
    \includegraphics[width=\linewidth]{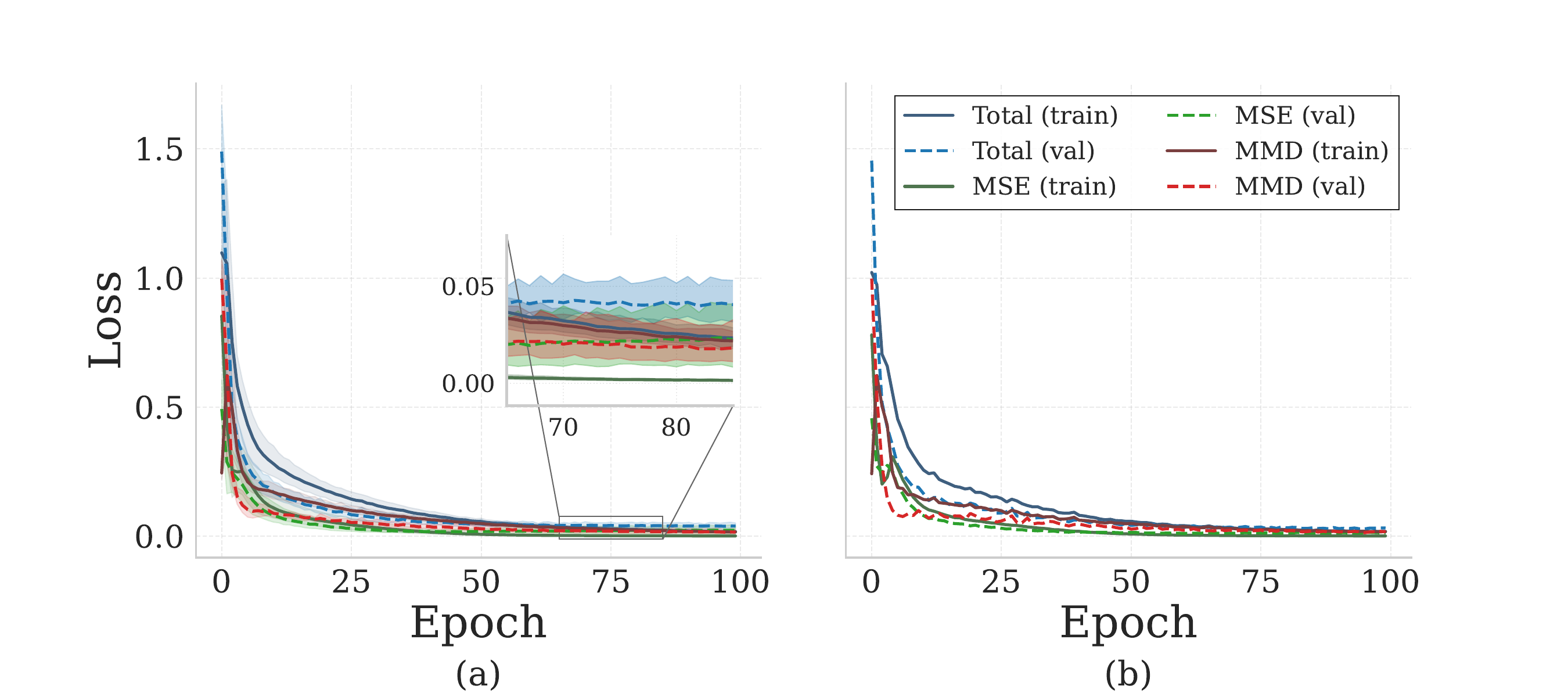}
    \caption{Training and validation loss evolution for the domain-shift aware model trained at the 1000 RPM regime with a balanced second-shaft condition. (a) Mean loss curves across 10 independent training runs, with shaded regions indicating the 5th–95th percentile range; the inset zooms into the late-training regime. (b) Loss curves of the best-performing run, defined as the run with the lowest final validation loss. Solid and dashed lines correspond to training and validation losses, respectively.}
    \label{fig:loss_1000_balanced}
\end{figure}

\paragraph{\textbf{(i) Training dynamics}}
The loss evolution in Fig.~\ref{fig:loss_1000_balanced} shows stable convergence across independent runs, with limited dispersion in the late-training regime. The small gap between training and validation losses indicates that overfitting is well controlled under the balanced second-shaft condition.

\begin{figure}[H]
    \centering
    \includegraphics[width=\linewidth]{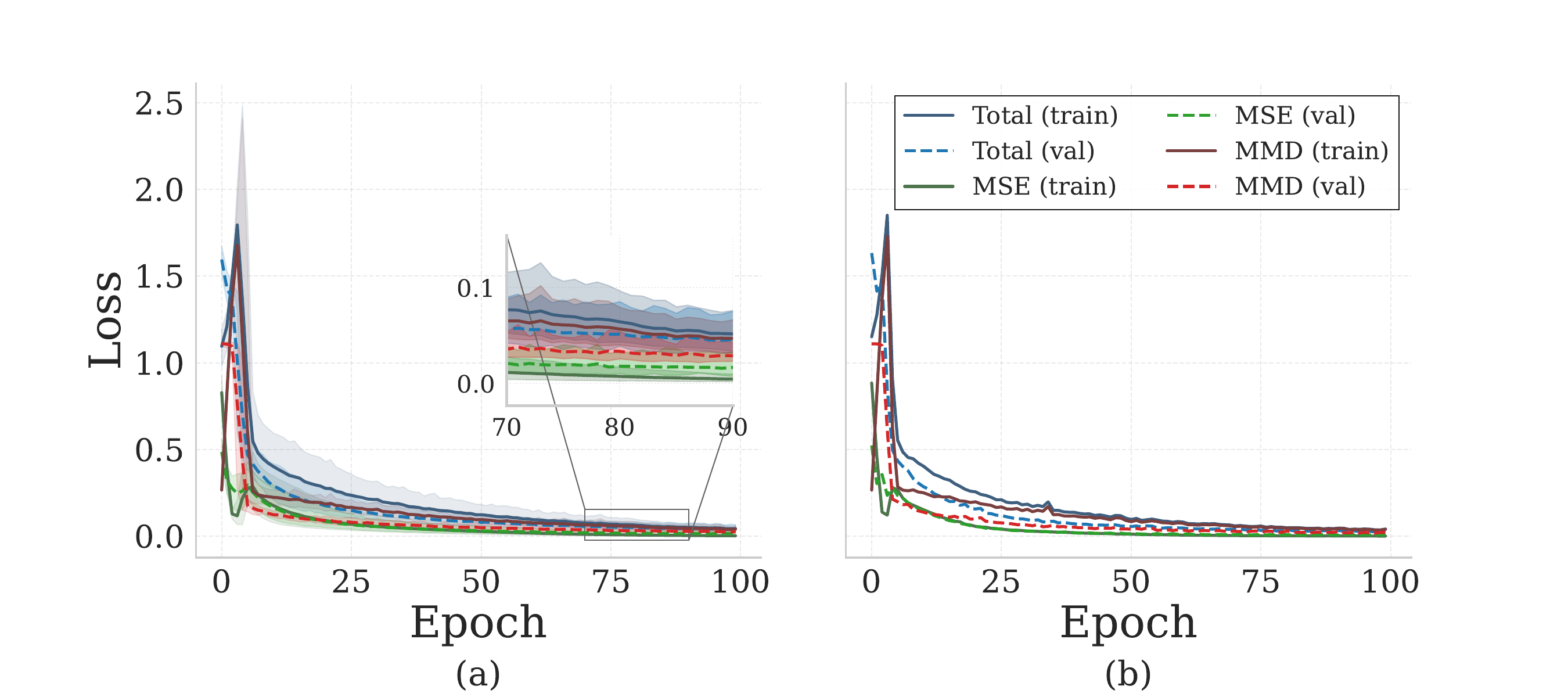}
    \caption{Training and validation loss evolution for the domain-shift aware model trained at the 1000 RPM regime with an unbalanced second-shaft condition. (a) Mean loss curves across 10 independent training runs, with shaded regions indicating the 5th–95th percentile range; the inset zooms into the late-training regime. (b) Loss curves of the best-performing run, defined as the run with the lowest final validation loss. Solid and dashed lines correspond to training and validation losses, respectively.}
    \label{fig:loss_1000_unbalanced}
\end{figure}

A similar trend is observed under the unbalanced second-shaft condition (Fig.~\ref{fig:loss_1000_unbalanced}), where convergence remains stable despite the increased variability. Although a slightly larger train–validation gap emerges, its magnitude remains limited, suggesting that the model retains good generalization even in the presence of higher structural discrepancy.

\begin{figure}[H]
    \centering
    \includegraphics[width=0.5\linewidth]{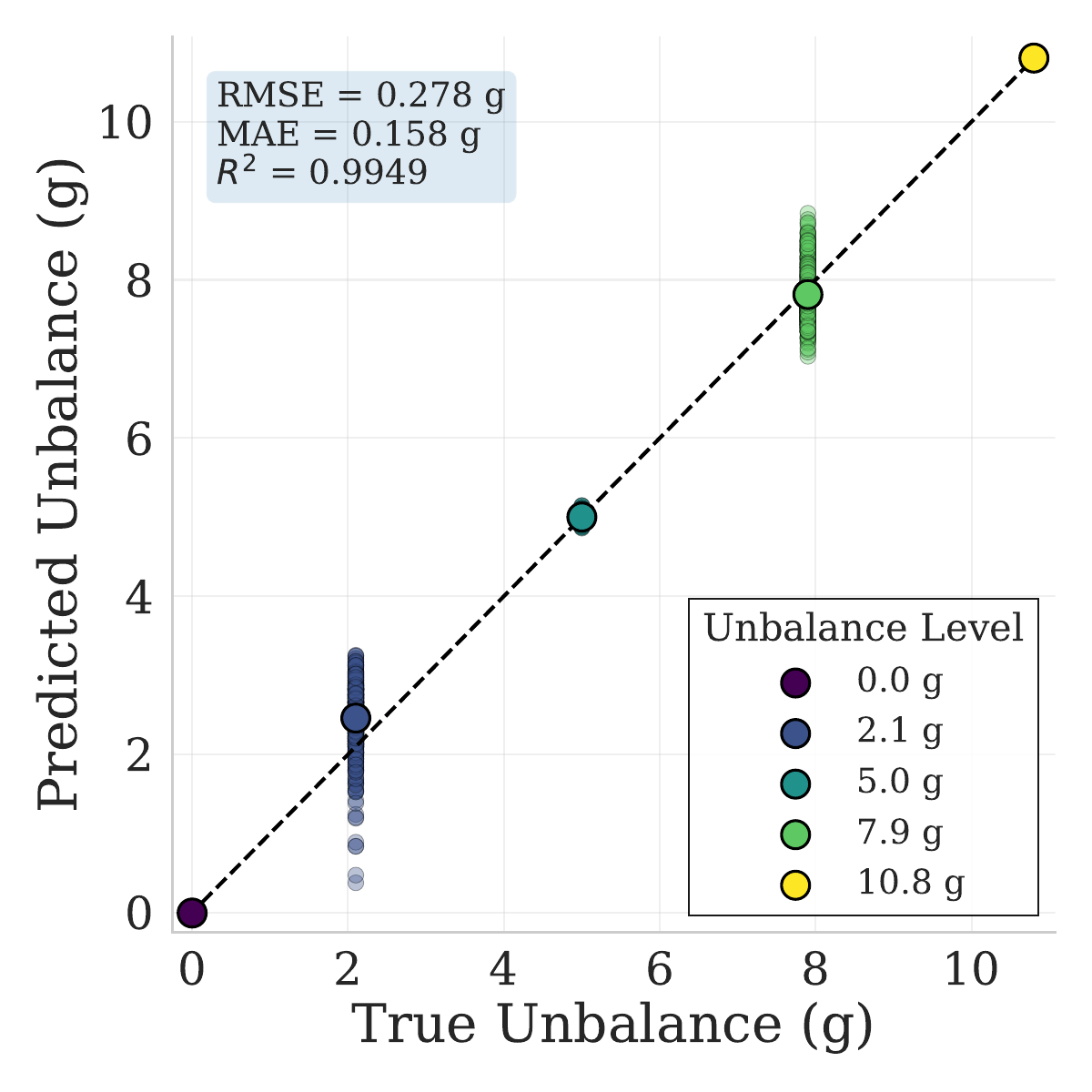}
    \caption{Predicted versus true unbalance values for the target domain at 1000 RPM under the balanced second-shaft condition using the domain-shift aware model. Each color represents a distinct unbalance level. Translucent markers indicate individual predictions, while larger markers denote the mean predicted value for each level. The dashed line represents ideal prediction. The dispersion of predictions around the ideal line reflects the estimation error for each unbalance level, with larger spread indicating lower accuracy and higher uncertainty. Systematic deviations of the mean predictions from the ideal line reveal bias in the model response. These effects directly impact the reported performance metrics, where increased dispersion and bias lead to higher RMSE and MAE values and a reduction in $R^2$. Consistent clustering along the ideal line indicates reliable generalization.}
    \label{fig:predictions_1000}
\end{figure}

\paragraph{\textbf{(ii) Predictive performance}}
This stable training behavior translates directly into strong predictive performance. Under the balanced second-shaft condition (Fig.~\ref{fig:predictions_1000}), predictions are tightly clustered around the ideal regression line, with minimal dispersion across all unbalance levels. The resulting metrics (RMSE = 0.278 g, MAE = 0.158 g, $R^2$ = 0.9949) indicate strong agreement with the ground truth.

\begin{figure}[H]
    \centering
    \includegraphics[width=0.5\linewidth]{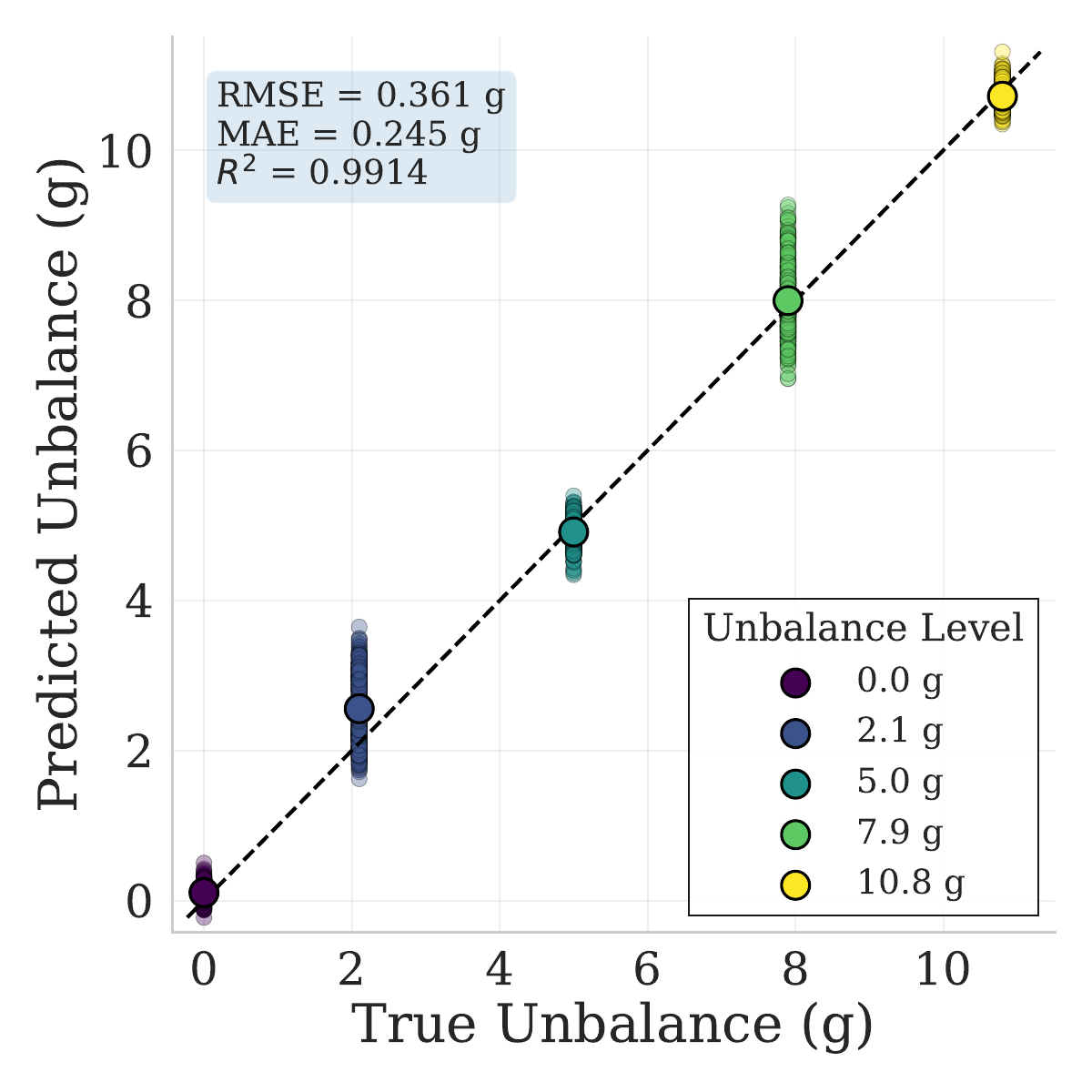}
    \caption{Predicted versus true unbalance values for the target domain at 1000 RPM under the unbalanced second-shaft condition using the domain-shift aware model. Each color represents a distinct unbalance level. Translucent markers indicate individual predictions, while larger markers denote the mean predicted value for each level. The dashed line represents ideal prediction. The dispersion of predictions around the ideal line reflects the estimation error for each unbalance level, with larger spread indicating lower accuracy and higher uncertainty. Systematic deviations of the mean predictions from the ideal line reveal bias in the model response. These effects directly impact the reported performance metrics, where increased dispersion and bias lead to higher RMSE and MAE values and a reduction in $R^2$. Consistent clustering along the ideal line indicates reliable generalization.}
    \label{fig:predictions_1000_unb}
\end{figure}

When higher structural discrepancy is introduced (Fig.~\ref{fig:predictions_1000_unb}), the model maintains high predictive consistency. A modest increase in dispersion is observed, but predictions remain well aligned with the ideal trend, as confirmed by the performance metrics (RMSE = 0.361 g, MAE = 0.245 g, $R^2$ = 0.9914).

\begin{figure}[H]
    \centering
    \includegraphics[width=0.6\linewidth]{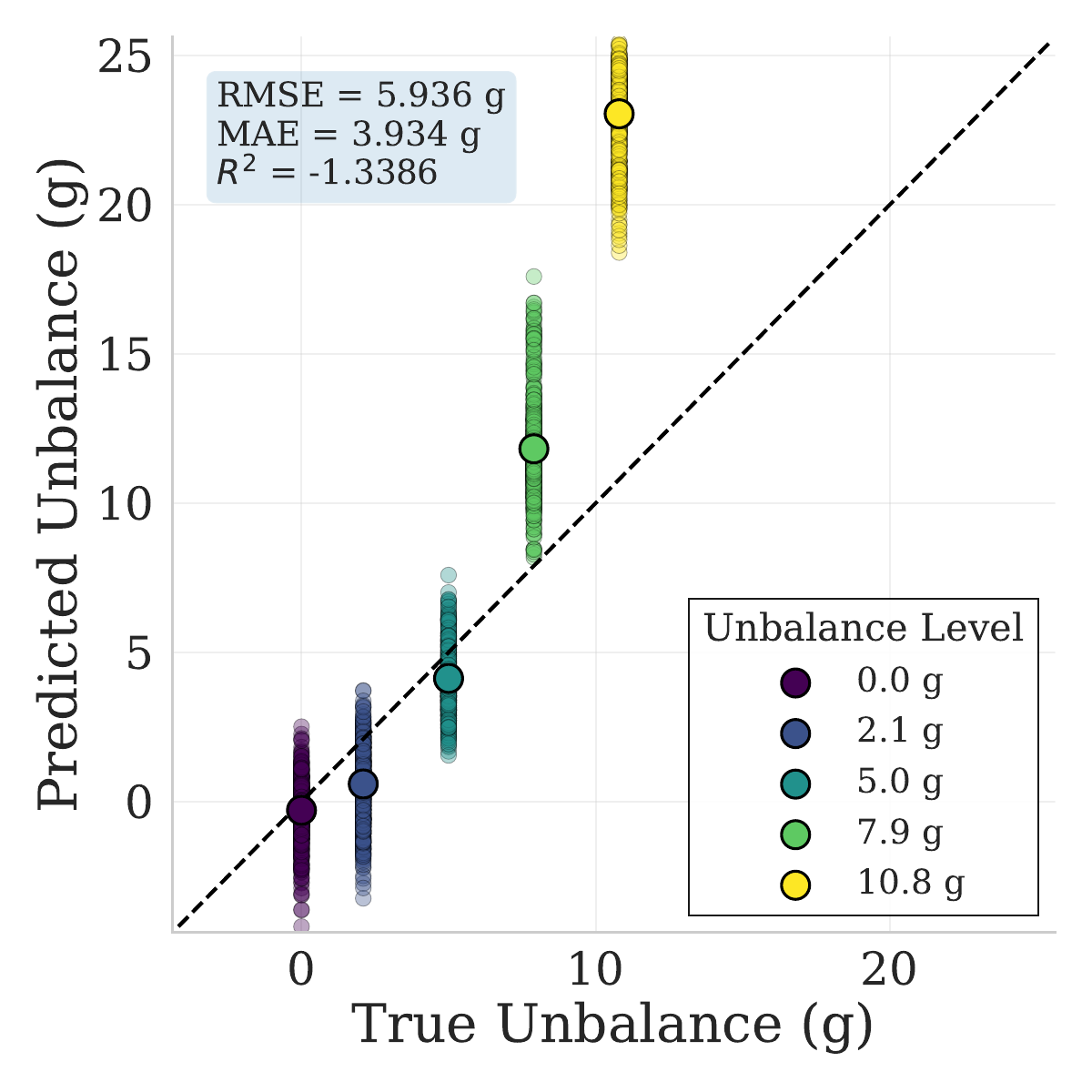}
    \caption{Predicted versus true unbalance values at 1000 RPM under the balanced second-shaft condition for the baseline model without domain alignment strategy (i.e., $\lambda = 0$ in Eq.~\ref{eq:total_loss}). Each color represents a distinct unbalance level. Translucent markers indicate individual predictions, while larger markers denote the mean predicted value for each level. The dashed line represents ideal prediction. The dispersion of predictions around the ideal line reflects the estimation error for each unbalance level, with larger spread indicating lower accuracy and higher uncertainty. Systematic deviations of the mean predictions from the ideal line reveal bias in the model response. These effects directly impact the reported performance metrics, where increased dispersion and bias lead to higher RMSE and MAE values and a reduction in $R^2$. Consistent clustering along the ideal line indicates reliable generalization.}
    \label{fig:predictions_1000_0}
\end{figure}

\begin{figure}[H]
    \centering
    \includegraphics[width=0.6\linewidth]{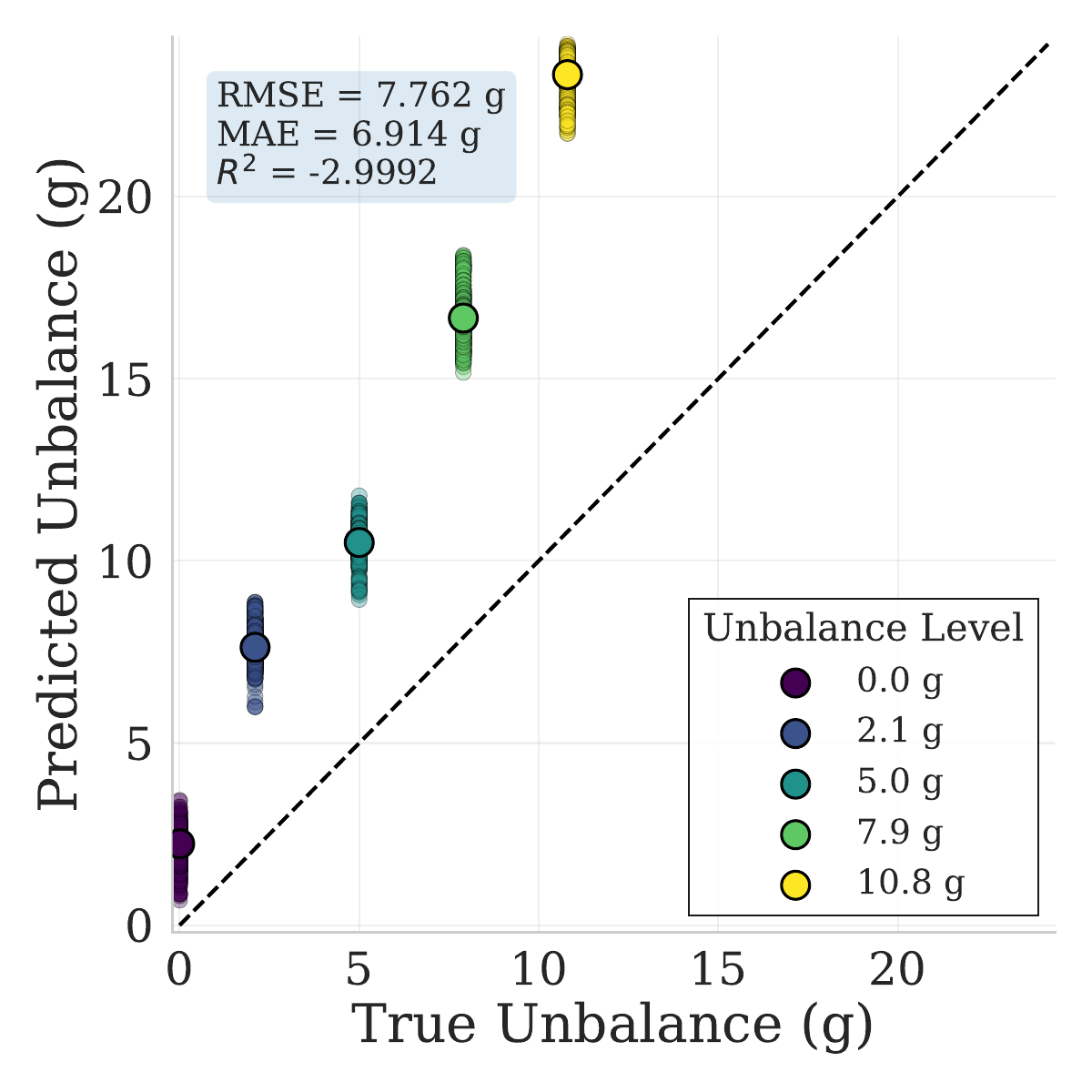}
    \caption{Predicted versus true unbalance values at 1000 RPM under the unbalanced second-shaft condition for the baseline model without domain alignment strategy (i.e., $\lambda = 0$ in Eq.~\ref{eq:total_loss}). Each color represents a distinct unbalance level. Translucent markers indicate individual predictions, while larger markers denote the mean predicted value for each level. The dashed line represents ideal prediction. The dispersion of predictions around the ideal line reflects the estimation error for each unbalance level, with larger spread indicating lower accuracy and higher uncertainty. Systematic deviations of the mean predictions from the ideal line reveal bias in the model response. These effects directly impact the reported performance metrics, where increased dispersion and bias lead to higher RMSE and MAE values and a reduction in $R^2$. Consistent clustering along the ideal line indicates reliable generalization.}
    \label{fig:predictions_1000_unb_0}
\end{figure}

The importance of domain alignment becomes evident when comparing with the baseline model. Figures~\ref{fig:predictions_1000_0} and \ref{fig:predictions_1000_unb_0} show that, without the MMD loss, predictions exhibit large dispersion and systematic bias, including physically inconsistent values. This leads to a substantial degradation in performance (e.g., RMSE = 5.936 g and negative $R^2$ in the balanced case), indicating a failure to generalize across domains.

\begin{figure}[H]
    \centering
    \includegraphics[width=\linewidth]{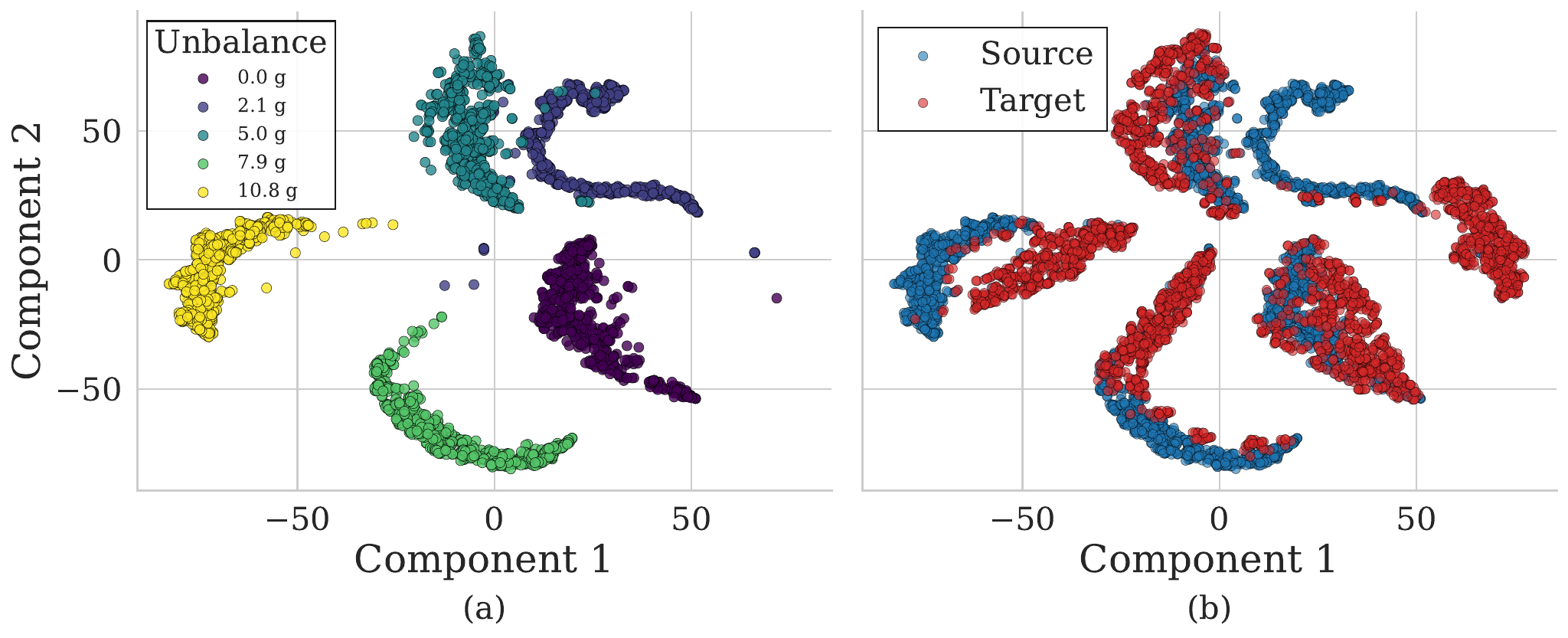}
    \caption{Two-dimensional t-SNE projection of the backbone feature representations extracted by the model at 1000 RPM under the balanced second-shaft condition. (a) Source-domain features, where each color represents a cluster associated with a specific unbalance level. Well-separated and compact clusters indicate high feature discriminability and facilitate reliable regression, including interpolation for unseen levels. In contrast, significant overlap between clusters reflects reduced separability, which can impair the model’s ability to distinguish between operating conditions and degrade predictive performance. (b) Embedded feature distributions of source and target domains, highlighting their spatial relationship in the learned feature space. A higher degree of overlap between domains suggests more effective domain alignment, provided that feature separability is preserved for accurate cross-domain regression.
    }
    \label{fig:tsne_1000}
\end{figure}

\paragraph{\textbf{(iii) Feature space structure}}
The t-SNE projections in Fig.~\ref{fig:tsne_1000} provide insight into the origin of this strong performance. In the source-domain embedding (Fig.~\ref{fig:tsne_1000}(a)), distinct clusters corresponding to different unbalance levels are clearly separated, indicating that the backbone extracts highly discriminative features.

In the joint embedding (Fig.~\ref{fig:tsne_1000}(b)), source and target samples exhibit substantial overlap, confirming that the domain alignment strategy effectively reduces distributional discrepancy while preserving discriminative structure.

\begin{figure}[H]
    \centering
    \includegraphics[width=\linewidth]{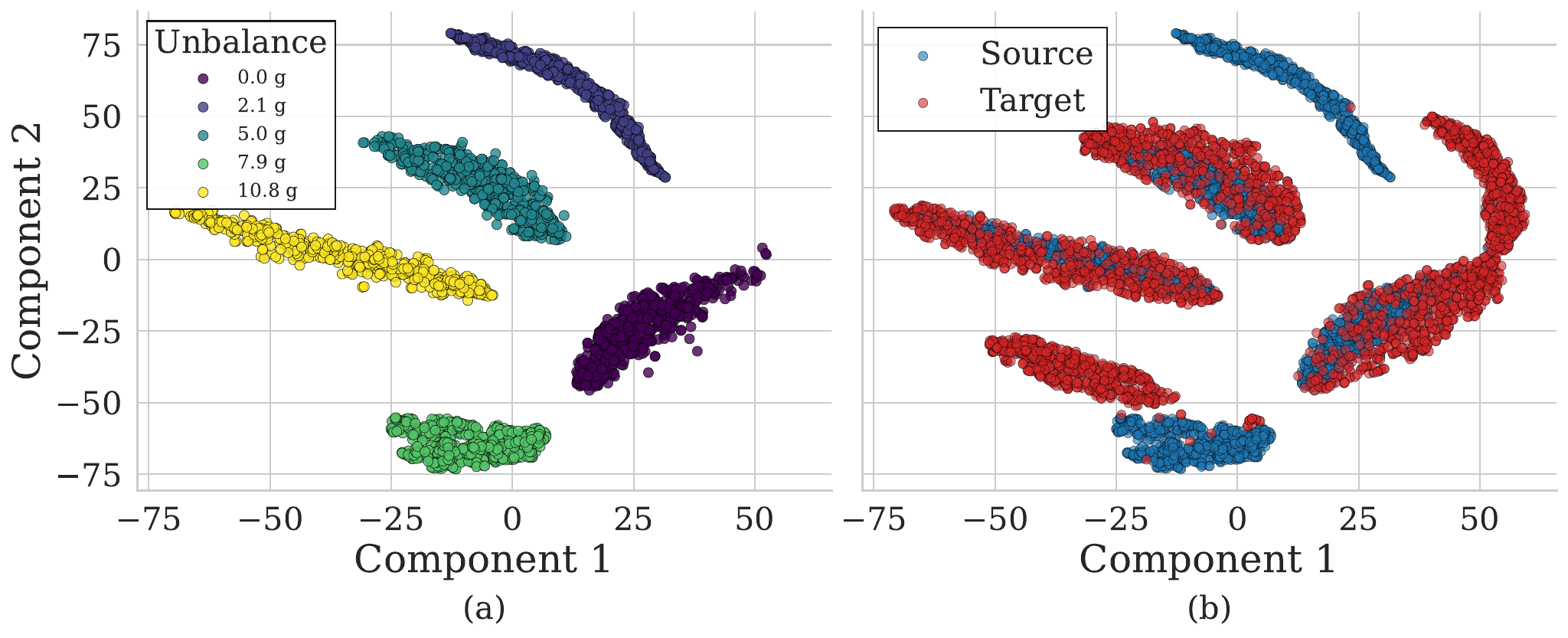}
    \caption{Two-dimensional t-SNE projection of the backbone feature representations extracted by the model at 1000 RPM under the unbalanced second-shaft condition. (a) Source-domain features, where each color represents a cluster associated with a specific unbalance level. Well-separated and compact clusters indicate high feature discriminability and facilitate reliable regression, including interpolation for unseen levels. In contrast, significant overlap between clusters reflects reduced separability, which can impair the model’s ability to distinguish between operating conditions and degrade predictive performance. (b) Embedded feature distributions of source and target domains, highlighting their spatial relationship in the learned feature space. A higher degree of overlap between domains suggests more effective domain alignment, provided that feature separability is preserved for accurate cross-domain regression.
    }
    \label{fig:tsne_1000_unb}
\end{figure}

A similar structure is observed under the unbalanced condition (Fig.~\ref{fig:tsne_1000_unb}), where clusters remain well separated and domain overlap is preserved. This indicates that the learned representation is both discriminative and transferable, even in the presence of increased structural variability.

Overall, the results at 1000 RPM demonstrate that effective feature separability combined with domain alignment strategy enables accurate and consistent cross-domain predictions. Unlike the low-speed regime, the data in this operating condition provide sufficient structure for the model to learn a discriminative and transferable representation.

\subsubsection{High-speed regime: 2000 RPM}

At 2000 RPM, the proposed model maintains strong generalization capability across domains, although the increased rotational speed introduces additional variability that slightly affects predictive performance. This regime highlights the impact of nonlinear vibration behavior and intensified domain discrepancies on cross-domain learning.

\begin{figure}[H]
    \centering
    \includegraphics[width=\linewidth]{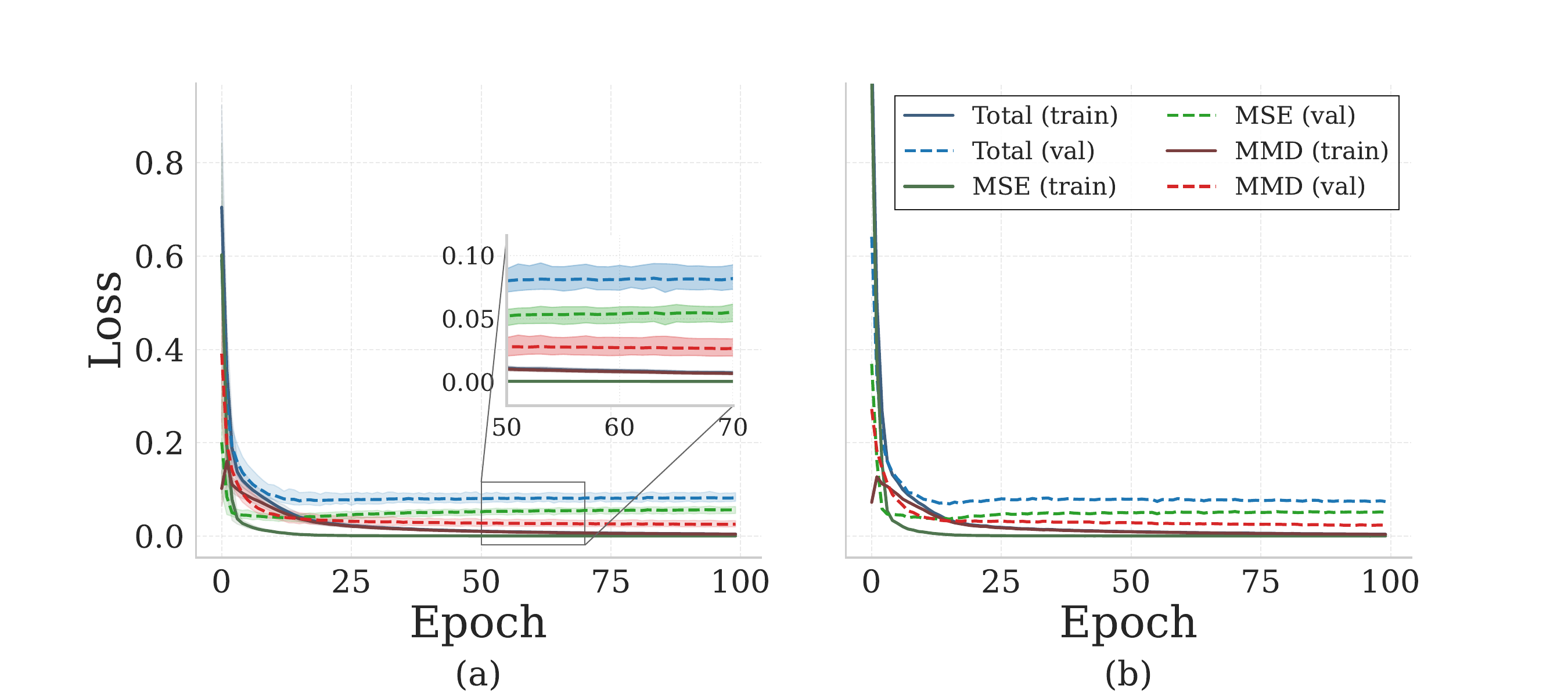}
    \caption{Training and validation loss evolution for the domain-shift aware model trained at the 2000 RPM regime with a balanced second-shaft condition. (a) Mean loss curves across 10 independent training runs, with shaded regions indicating the 5th–95th percentile range; the inset zooms into the late-training regime. (b) Loss curves of the best-performing run, defined as the run with the lowest final validation loss. Solid and dashed lines correspond to training and validation losses, respectively.}
    \label{fig:loss_2000_balanced}
\end{figure}

\paragraph{\textbf{(i) Training dynamics}}
The loss evolution in Fig.~\ref{fig:loss_2000_balanced} shows stable convergence across independent runs, indicating consistent optimization dynamics. However, compared to the 1000 RPM case, a larger dispersion is observed in the late-training regime, along with a slightly wider gap between training and validation losses, suggesting a moderate increase in overfitting.

\begin{figure}[H]
    \centering
    \includegraphics[width=\linewidth]{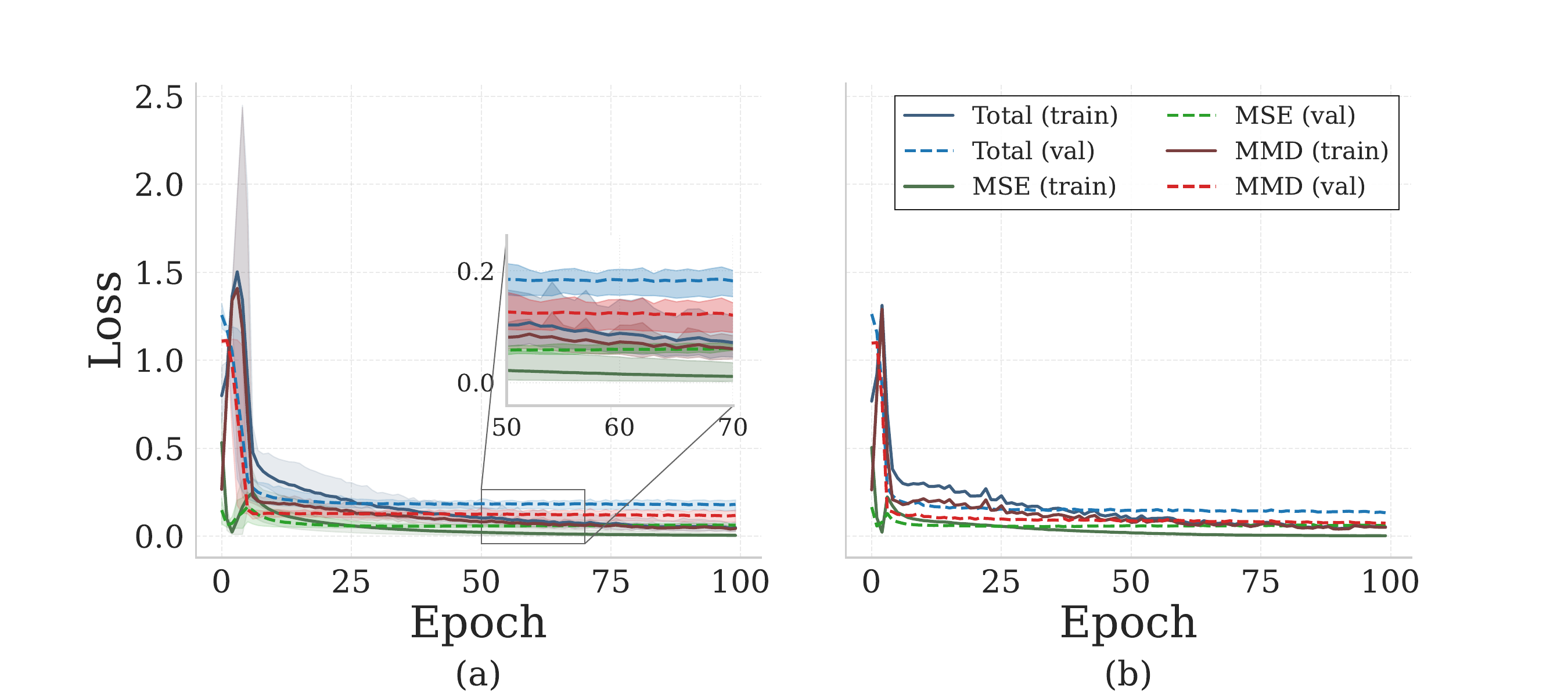}
    \caption{Training and validation loss evolution for the domain-shift aware model trained at the 2000 RPM regime with an unbalanced second-shaft condition. (a) Mean loss curves across 10 independent training runs, with shaded regions indicating the 5th–95th percentile range; the inset zooms into the late-training regime. (b) Loss curves of the best-performing run, defined as the run with the lowest final validation loss. Solid and dashed lines correspond to training and validation losses, respectively.}
    \label{fig:loss_2000_unbalanced}
\end{figure}

Under the unbalanced second-shaft condition (Fig.~\ref{fig:loss_2000_unbalanced}), this effect becomes more pronounced, with increased variability across runs and a clearer separation between training and validation curves, indicating reduced training stability and higher sensitivity to structural discrepancy.

\begin{figure}[H]
    \centering
    \includegraphics[width=0.6\linewidth]{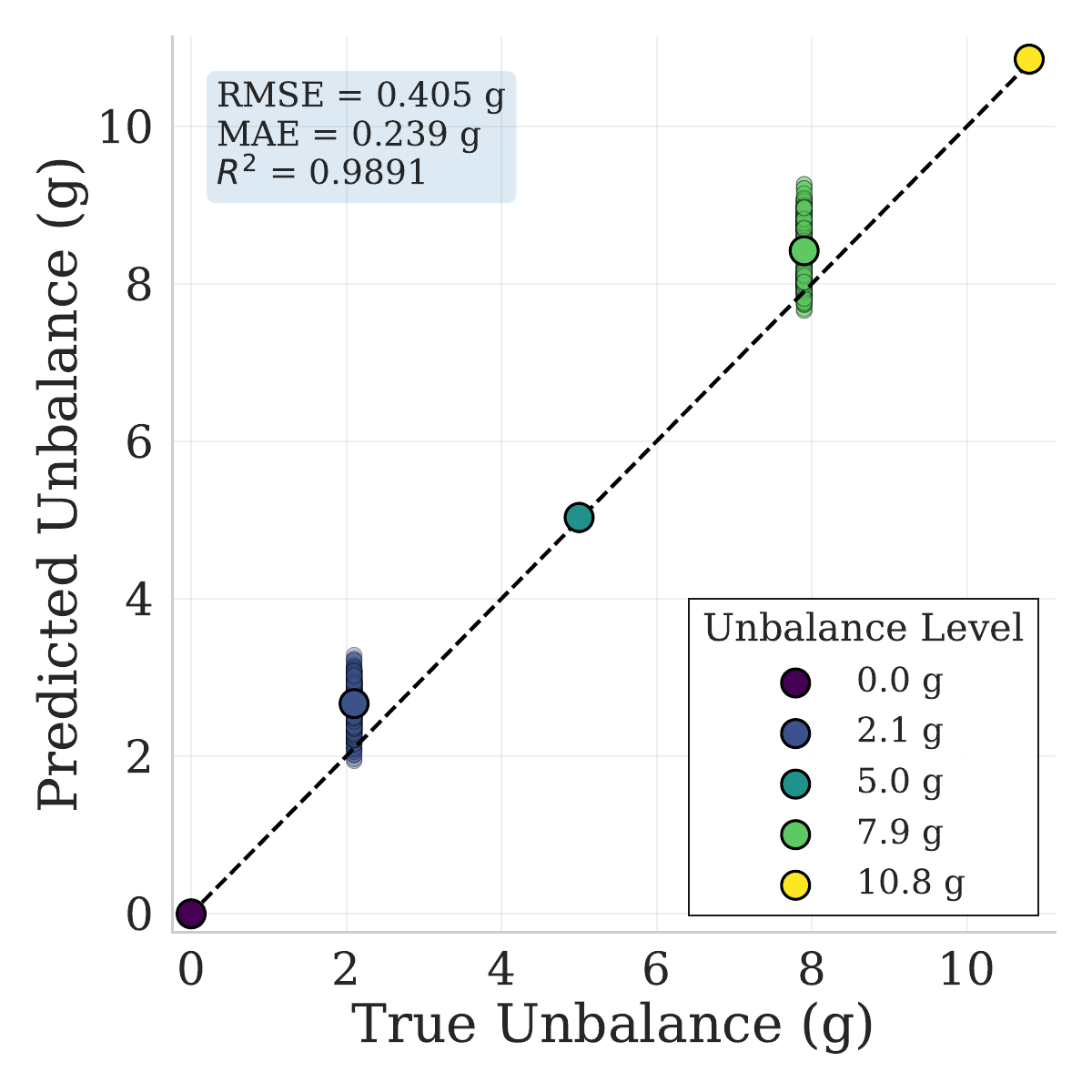}
    \caption{Predicted versus true unbalance values for the target domain at 2000 RPM under the balanced second-shaft condition using the domain-shift aware model. Each color represents a distinct unbalance level. Translucent markers indicate individual predictions, while larger markers denote the mean predicted value for each level. The dashed line represents ideal prediction. The dispersion of predictions around the ideal line reflects the estimation error for each unbalance level, with larger spread indicating lower accuracy and higher uncertainty. Systematic deviations of the mean predictions from the ideal line reveal bias in the model response. These effects directly impact the reported performance metrics, where increased dispersion and bias lead to higher RMSE and MAE values and a reduction in $R^2$. Consistent clustering along the ideal line indicates reliable generalization.}
    \label{fig:predictions_2000}
\end{figure}

\paragraph{\textbf{(ii) Predictive performance}}
These effects are reflected in the predictive results. Under the balanced second-shaft condition (Fig.~\ref{fig:predictions_2000}), the model exhibits strong agreement with the ground truth, with predictions closely distributed around the ideal regression line. A slight increase in dispersion is observed compared to the 1000 RPM case, particularly for unbalance levels not present during training, leading to a moderate degradation in performance (RMSE = 0.405 g, MAE = 0.239 g, $R^2$ = 0.9891).

\begin{figure}[H]
    \centering
    \includegraphics[width=0.6\linewidth]{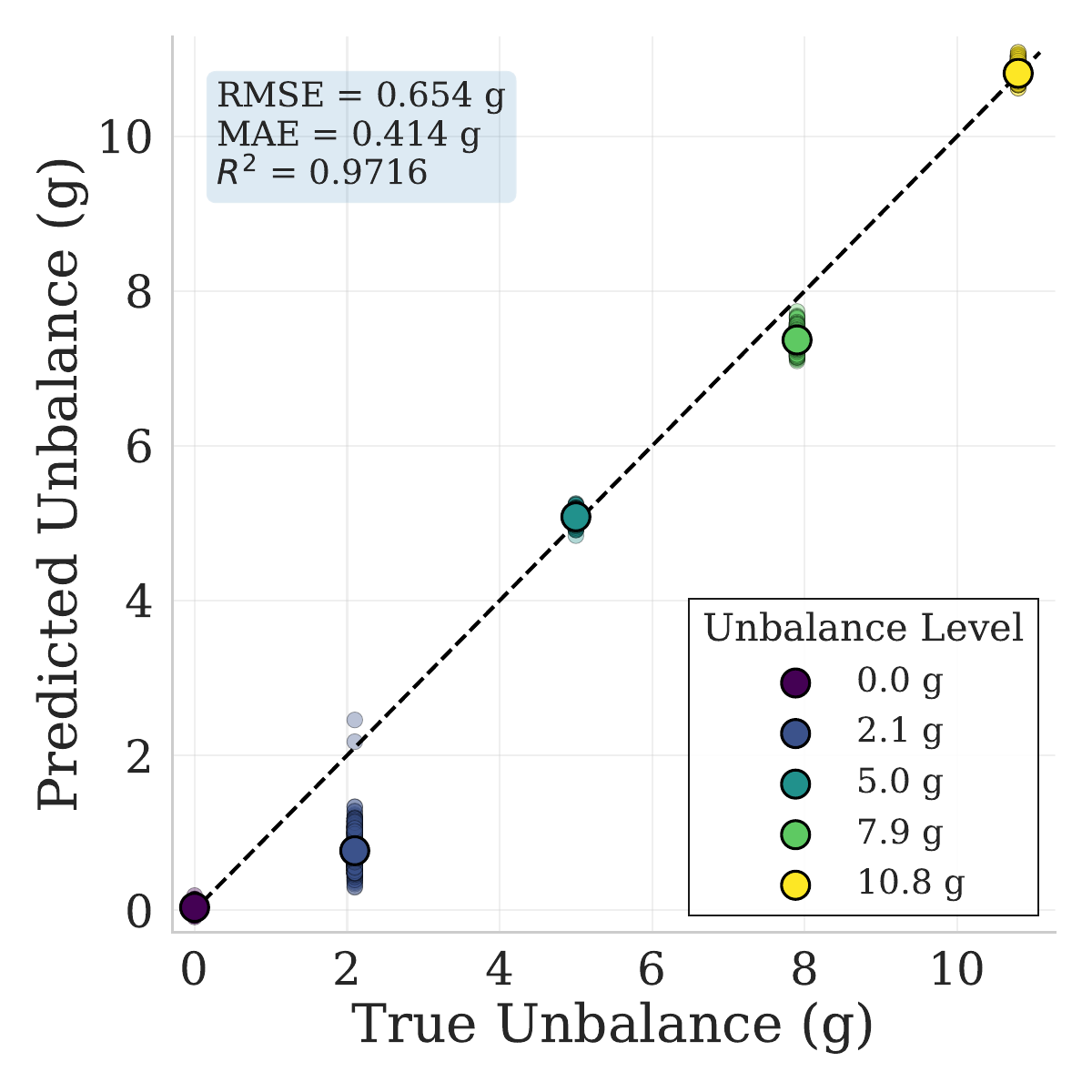}
    \caption{Predicted versus true unbalance values for the target domain at 2000 RPM under the unbalanced second-shaft condition using the domain-shift aware model. Each color represents a distinct unbalance level. Translucent markers indicate individual predictions, while larger markers denote the mean predicted value for each level. The dashed line represents ideal prediction. The dispersion of predictions around the ideal line reflects the estimation error for each unbalance level, with larger spread indicating lower accuracy and higher uncertainty. Systematic deviations of the mean predictions from the ideal line reveal bias in the model response. These effects directly impact the reported performance metrics, where increased dispersion and bias lead to higher RMSE and MAE values and a reduction in $R^2$. Consistent clustering along the ideal line indicates reliable generalization.}
    \label{fig:predictions_2000_unb}
\end{figure}

When higher structural discrepancy is introduced (Fig.~\ref{fig:predictions_2000_unb}), the degradation becomes more evident. Although the model still captures the overall trend, increased dispersion and bias are observed, particularly at lower unbalance levels, resulting in reduced accuracy (RMSE = 0.654 g, MAE = 0.414 g, $R^2$ = 0.9716).

\begin{figure}[H]
    \centering
    \includegraphics[width=0.6\linewidth]{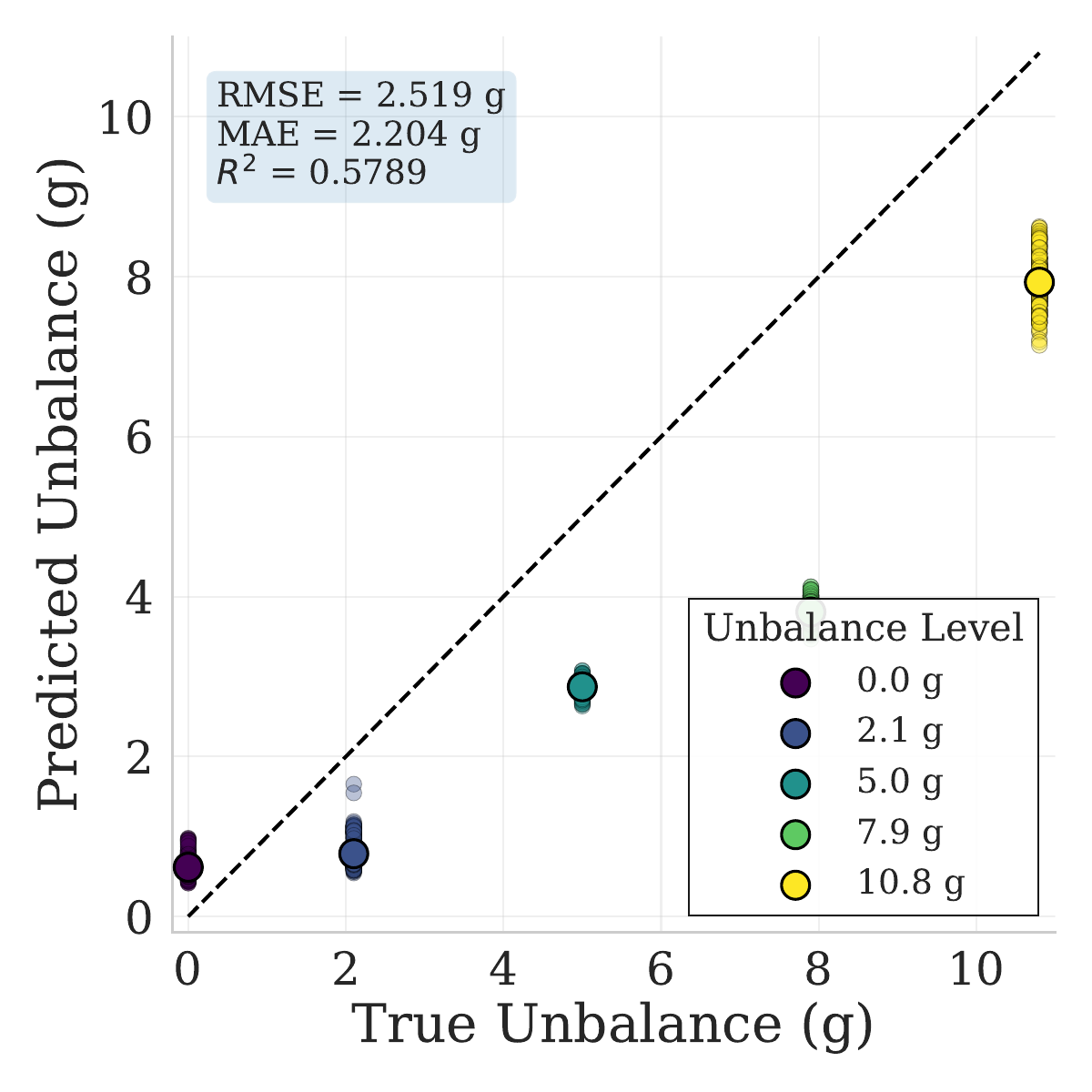}
    \caption{Predicted versus true unbalance values at 2000 RPM under the balanced second-shaft condition for the baseline model without domain alignment strategy (i.e., $\lambda = 0$ in Eq.~\ref{eq:total_loss}). Each color represents a distinct unbalance level. Translucent markers indicate individual predictions, while larger markers denote the mean predicted value for each level. The dashed line represents ideal prediction. The dispersion of predictions around the ideal line reflects the estimation error for each unbalance level, with larger spread indicating lower accuracy and higher uncertainty. Systematic deviations of the mean predictions from the ideal line reveal bias in the model response. These effects directly impact the reported performance metrics, where increased dispersion and bias lead to higher RMSE and MAE values and a reduction in $R^2$. Consistent clustering along the ideal line indicates reliable generalization.}
    \label{fig:predictions_2000_0}
\end{figure}

\begin{figure}[H]
    \centering
    \includegraphics[width=0.6\linewidth]{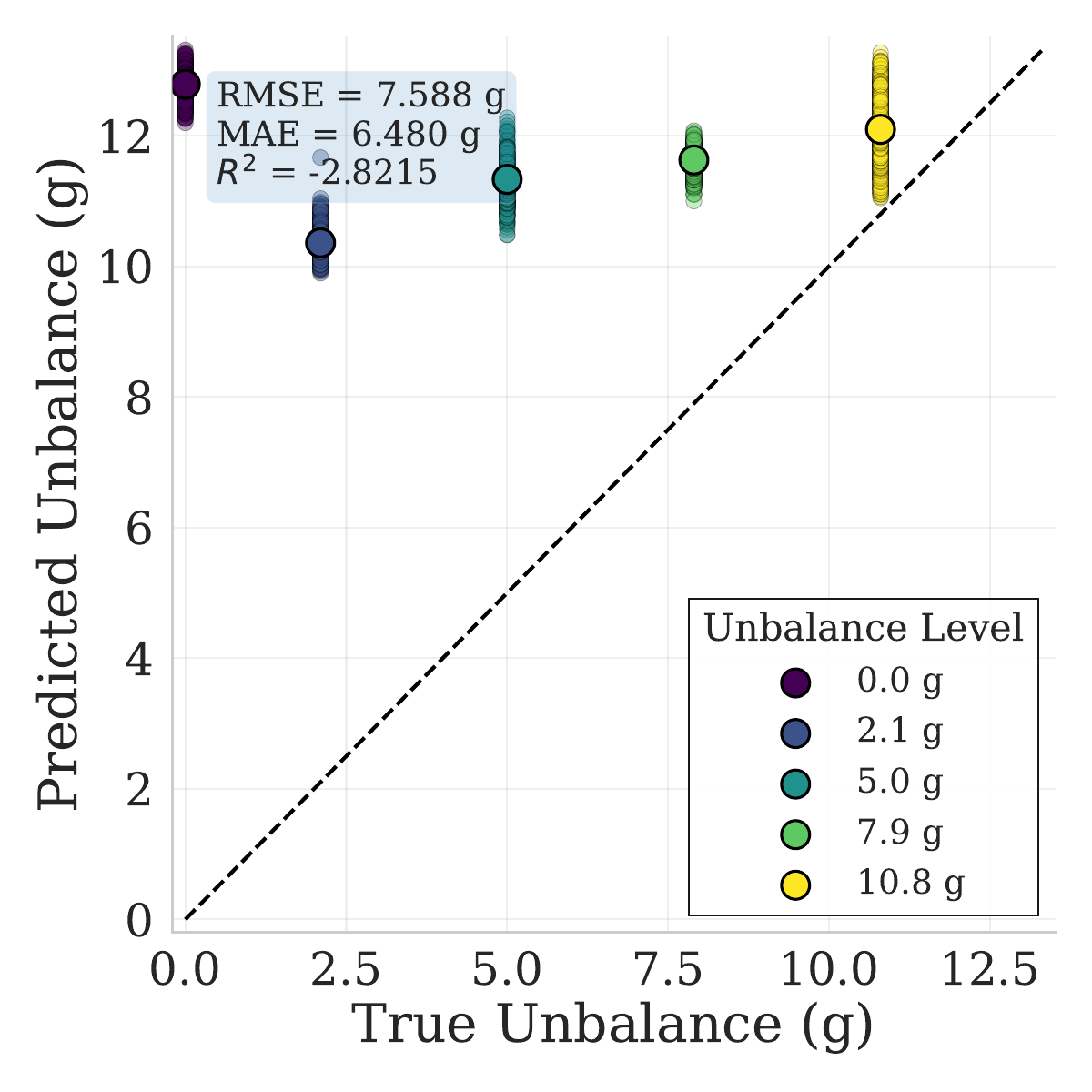}
    \caption{Predicted versus true unbalance values at 2000 RPM under the unbalanced second-shaft condition for the baseline model without domain alignment strategy (i.e., $\lambda = 0$ in Eq.~\ref{eq:total_loss}). Each color represents a distinct unbalance level. Translucent markers indicate individual predictions, while larger markers denote the mean predicted value for each level. The dashed line represents ideal prediction. The dispersion of predictions around the ideal line reflects the estimation error for each unbalance level, with larger spread indicating lower accuracy and higher uncertainty. Systematic deviations of the mean predictions from the ideal line reveal bias in the model response. These effects directly impact the reported performance metrics, where increased dispersion and bias lead to higher RMSE and MAE values and a reduction in $R^2$. Consistent clustering along the ideal line indicates reliable generalization.}
    \label{fig:predictions_2000_unb_0}
\end{figure}

The comparison with the baseline model highlights the importance of domain alignment. Figures~\ref{fig:predictions_2000_0} and \ref{fig:predictions_2000_unb_0} show that, without the MMD loss, predictions become highly biased and dispersed, leading to a severe degradation in performance and strongly negative $R^2$ values.

\begin{figure}[H]
    \centering
    \includegraphics[width=\linewidth]{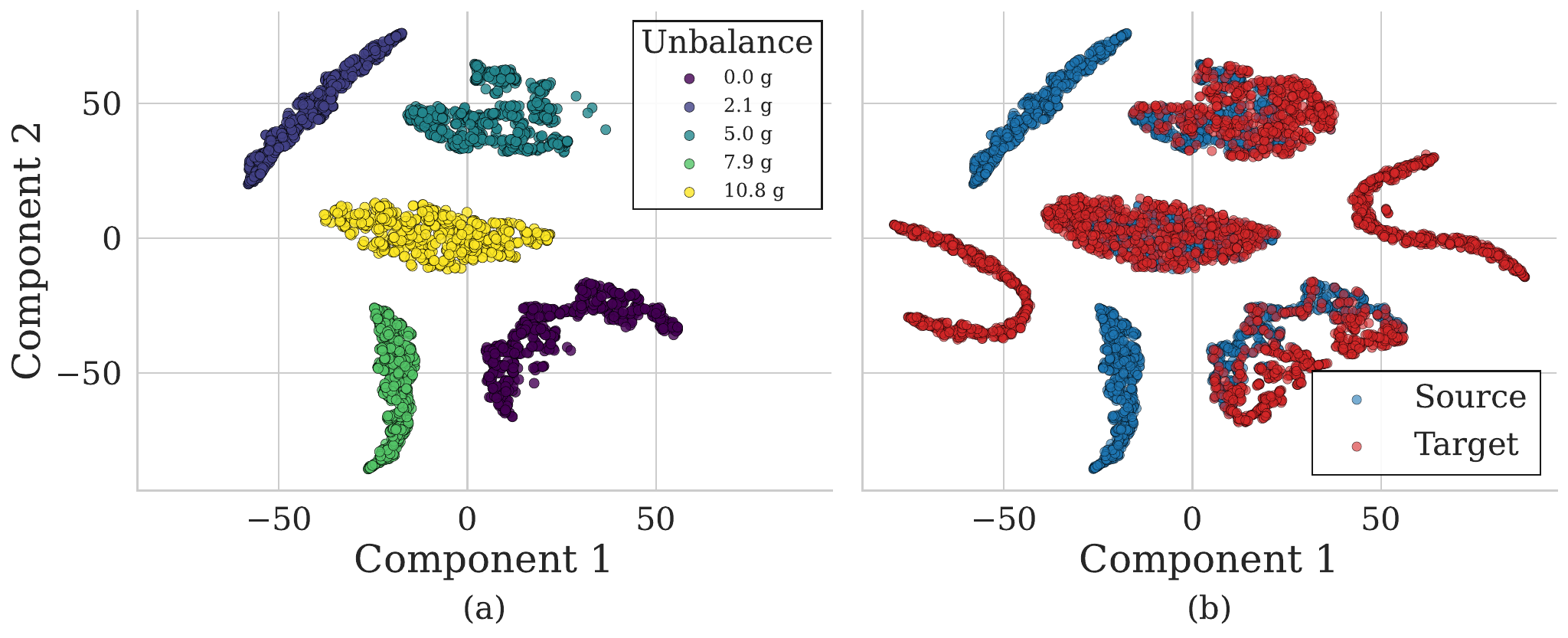}
    \caption{Two-dimensional t-SNE projection of the backbone feature representations extracted by the model at 2000 RPM under the balanced second-shaft condition. (a) Source-domain features, where each color represents a cluster associated with a specific unbalance level. Well-separated and compact clusters indicate high feature discriminability and facilitate reliable regression, including interpolation for unseen levels. In contrast, significant overlap between clusters reflects reduced separability, which can impair the model’s ability to distinguish between operating conditions and degrade predictive performance. (b) Embedded feature distributions of source and target domains, highlighting their spatial relationship in the learned feature space. A higher degree of overlap between domains suggests more effective domain alignment, provided that feature separability is preserved for accurate cross-domain regression.
    }
    \label{fig:tsne_2000}
\end{figure}

\paragraph{\textbf{(iii) Feature space structure}}
The t-SNE projections in Fig.~\ref{fig:tsne_2000} provide further insight into these behaviors. In the source-domain embedding (Fig.~\ref{fig:tsne_2000}(a)), distinct clusters associated with different unbalance levels remain well separated, indicating that the learned representation preserves discriminative information even at higher rotational speeds. In the joint embedding (Fig.~\ref{fig:tsne_2000}(b)), source and target domains largely overlap, confirming the effectiveness of the domain alignment strategy.

However, compared to the 1000 RPM case, the alignment appears less compact, particularly for unbalance levels not seen during training, where a slight separation between domains emerges. This behavior is consistent with the observed increase in prediction error.

\begin{figure}[H]
    \centering
    \includegraphics[width=\linewidth]{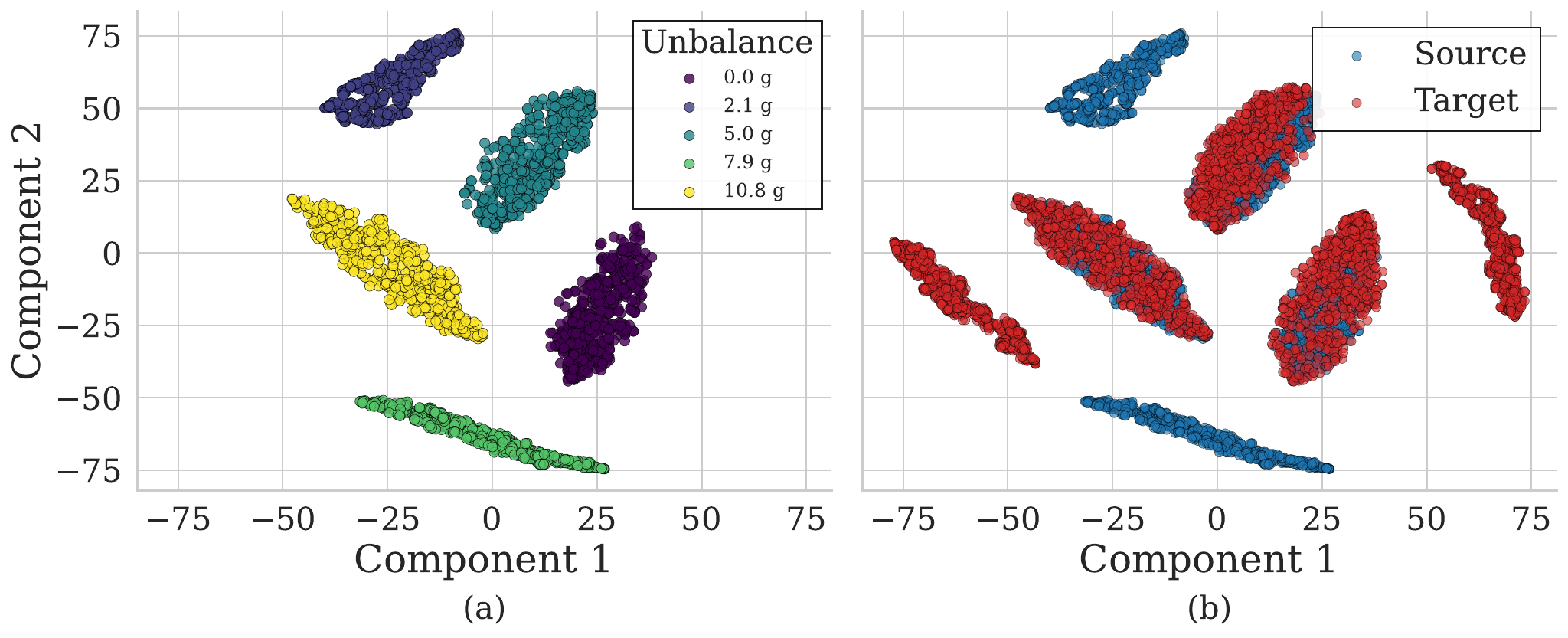}
    \caption{Two-dimensional t-SNE projection of the backbone feature representations extracted by the model at 2000 RPM under the unbalanced second-shaft condition. (a) Source-domain features, where each color represents a cluster associated with a specific unbalance level. Well-separated and compact clusters indicate high feature discriminability and facilitate reliable regression, including interpolation for unseen levels. In contrast, significant overlap between clusters reflects reduced separability, which can impair the model’s ability to distinguish between operating conditions and degrade predictive performance. (b) Embedded feature distributions of source and target domains, highlighting their spatial relationship in the learned feature space. A higher degree of overlap between domains suggests more effective domain alignment, provided that feature separability is preserved for accurate cross-domain regression.
    }
    \label{fig:tsne_2000_unb}
\end{figure}

Under the unbalanced condition (Fig.~\ref{fig:tsne_2000_unb}), feature distributions become more dispersed and partially misaligned, especially for unseen operating conditions, reflecting the increased difficulty of simultaneously handling domain shift and higher structural discrepancy.

Overall, the results at 2000 RPM indicate that, although domain alignment remains effective, increased operational variability and structural discrepancy introduce additional challenges that limit feature compactness and slightly degrade predictive accuracy.

\subsection{Discussion}\label{subsec4-1}

The analysis of the three operating conditions indicates that the best overall performance was achieved at 1000 RPM, followed by 2000 RPM, while 500 RPM yielded the poorest results. In the latter case, some predictions lacked physical meaning (e.g., negative unbalance mass estimates), indicating insufficient dynamic excitation.

Considering that the unbalance mass is located at a radial distance of 40 mm from the shaft center, the centrifugal force can be approximately estimated as $F \approx 0.04 \ m \ \omega^2$, where $m$ is the unbalance mass in kilograms and $\omega$ is the rotational speed in rad/s. At 500 RPM, the largest tested mass (10.8 g) generates a centrifugal force of approximately 1.18 N, while at 1000 RPM the smallest mass (2.1 g) produces approximately 0.92 N. Although the absolute force levels are of the same order of magnitude, the higher rotational speed increases the excitation frequency, leading to a more energetic and distinguishable dynamic response. This suggests that the improved performance at 1000 RPM is not solely related to force magnitude, but rather to a more favorable dynamic excitation regime, enhancing signal separability. This effect is further evidenced in the frequency-domain representations discussed next.

\begin{figure}[H]
\centering
\includegraphics[width=0.5\linewidth]{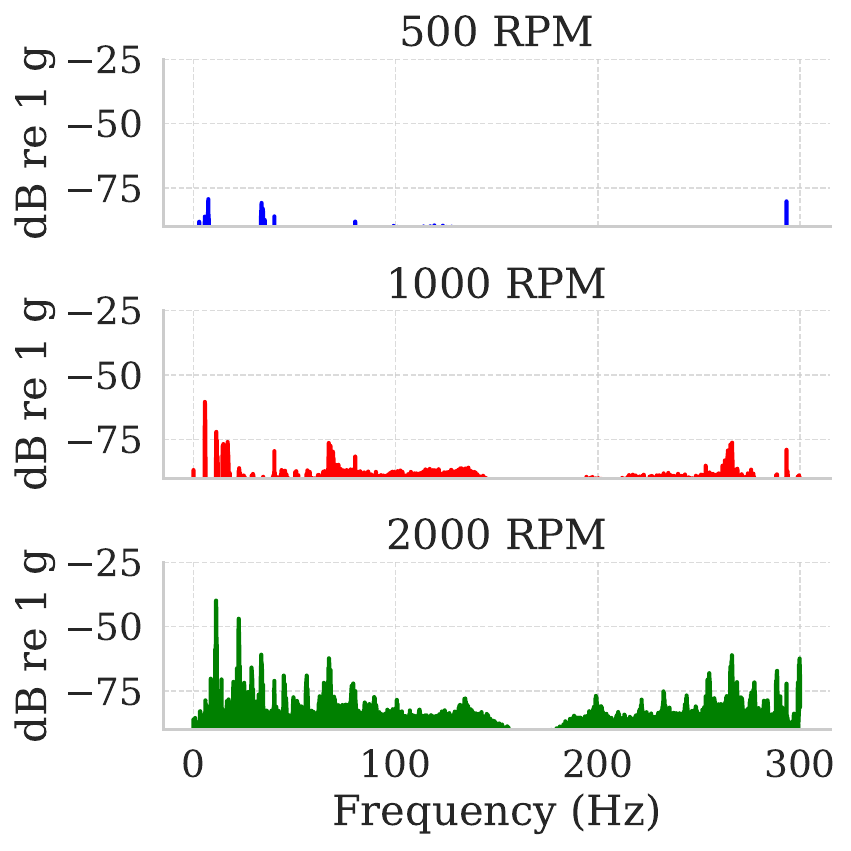}
\caption{Representative frequency spectra for 500, 1000, and 2000 RPM for a sample with 5.0 g of unbalanced mass. The amplitude axis is shown in decibels relative to 1 g (dB re 1 g).}
\label{fig:fft_comparison}
\end{figure}

Figure~\ref{fig:fft_comparison} presents representative frequency spectra for the three operating speeds for a sample with an unbalanced mass of 5.0 g. The spectral amplitude is presented in decibels relative to 1 g (dB re 1 g), providing a logarithmic representation of the vibration acceleration referenced to gravitational acceleration. A clear increase in spectral amplitude with rotational speed can be observed. At 500 RPM, the low excitation level results in weak spectral peaks, making the dynamic signature less distinguishable. Consistent with the previously discussed excitation characteristics, particularly the reduced centrifugal forces at lower rotational speeds, the vibration response becomes more susceptible to background noise and measurement variability, leading to a reduced signal-to-noise ratio (SNR) and poorer separability of characteristic frequency components. This directly impacts the ability of the CNN to extract discriminative features, as the learned convolutional filters rely on consistent and salient patterns in the input signals. At 1000 RPM, the dominant frequency component becomes more pronounced, improving signal separability. At 2000 RPM, although the spectral amplitude increases significantly, additional harmonic components and broader spectral content emerge, suggesting increased dynamic complexity. 

At 2000 RPM, despite the substantially higher centrifugal force (approximately 3.68 N for 2.1 g), a slight degradation in performance was observed. This behavior may be associated with increased nonlinear effects and additional dynamic complexities at higher rotational speeds \cite{randall2021}, which can reduce model generalization. Moreover, because the unbalance mass is located outside the flange, it generates an additional torque on the shaft, potentially amplifying nonlinear interactions and contributing to the observed increase in dynamic complexity.

It is important to note that these findings are specific to the experimental setup and operating conditions considered, and should be interpreted within this context.

\section{Conclusions}\label{sec5}

This work proposed and analyzed a domain-shift aware neural network for estimating unbalance masses in a rotating system with domain discrepancies. Experimental data were collected from a test rig, where a primary shaft coupled to an electric motor was equipped with a flange carrying unbalanced masses, while a secondary shaft, driven by a second motor, could also be activated to introduce domain discrepancy. Within this context, the work serves as a case study to introduce and evaluate the feasibility of applying domain-adaptation strategies to regression problems in SHM.

Within the considered experimental conditions, the model achieved good performance at medium (1000 RPM) and high (2000 RPM) rotational speeds, with the best results observed at medium speeds. At low speeds (500 RPM), performance was limited, as the low excitation levels reduced the dynamic signature associated with the unbalance.

Performance degradation was consistently observed when the second shaft was also unbalanced, as expected, since the domain discrepancy increases under this condition. As the difference between source and target distributions becomes more pronounced, aligning feature representations through the MMD-based domain adaptation strategy becomes more challenging.

Importantly, the domain-shift aware model significantly outperformed conventional models of identical architecture trained solely with a data-fidelity loss function, demonstrating the effectiveness of explicitly addressing distribution mismatch. Although the results demonstrate the potential of the proposed approach, they are inherently dependent on the characteristics of the dataset, including noise levels, excitation regimes, and the specific mechanical configuration of the test rig. Therefore, the conclusions should not be interpreted as universally generalizable, but rather as evidence supporting the feasibility and potential of domain-adaptive regression strategies in controlled SHM scenarios.

In the present work, unbalance masses were always positioned at a fixed radial distance. Future work could explore varying this distance as an additional source of domain discrepancy, since it directly affects the system’s dynamic response. Moreover, integrating physics-informed modeling strategies with domain adaptation mechanisms may further improve robustness and generalization, particularly under severe distribution shifts. Finally, extending the proposed approach to other rotating systems would enable a broader evaluation of the generalization capabilities of domain-adaptive regression models in SHM applications.

\backmatter

\bmhead{Acknowledgements}

Daniel Alves Castello would like to express his gratitude for the financial support provided by CNPq under grant number 312669/2023-2.

\section*{Declarations}

\subsection*{Funding}
Not applicable.

\subsection*{Consent for publication}
Not applicable.

\subsection*{Data availability}
The dataset generated and analyzed during this work is publicly available in the GitHub repository "Characterization of Unbalance in Rotating Systems", archived at Zenodo (\url{https://doi.org/10.5281/zenodo.18775036}).

\subsection*{Author contributions}

\noindent \textbf{Bernardo Feijó Junqueira:} Conceptualization, Methodology, Software, Validation, Formal analysis, Investigation, Visualization, Supervision, Writing – original draft, Writing – review \& editing.

\noindent \textbf{Claudio Kiyoshi Umezu:} Conceptualization, Methodology, Investigation, Data curation, Supervision, Writing – original draft, Writing – review \& editing.

\noindent \textbf{Bruno Bilhar Karaziack:} Conceptualization, Methodology, Validation, Investigation, Data curation, Writing – original draft, Writing – review \& editing.

\noindent \textbf{Tomaz Junior:} Conceptualization, Methodology, Validation, Investigation, Data curation, Writing – review \& editing.

\noindent \textbf{Daniel Alves Castello:} Methodology, Supervision, Writing – review \& editing.

\begin{appendices}

\section{Dataset Description}\label{secA1}

\begin{figure}[h]
    \centering
    \includegraphics[width=0.8\linewidth]{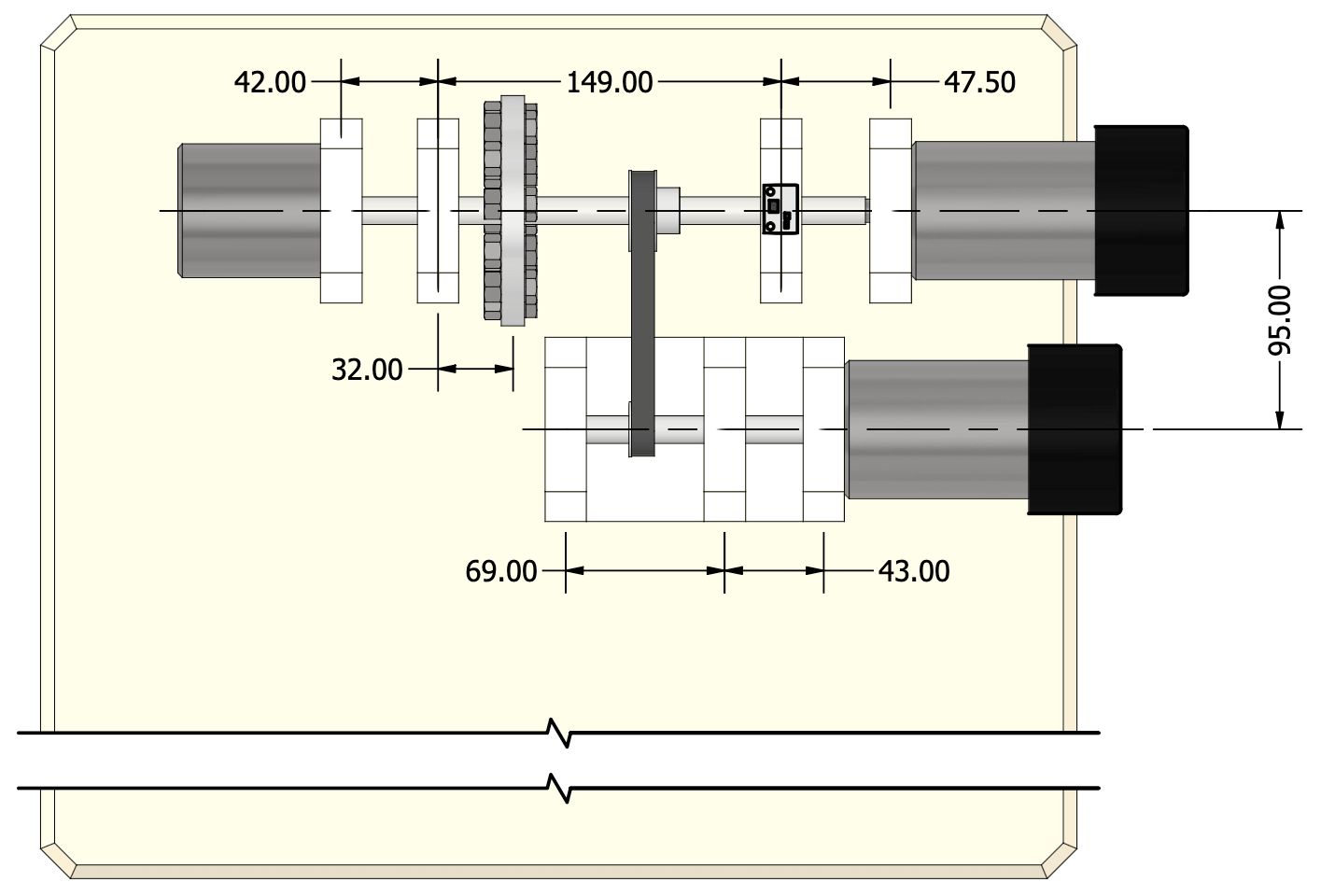}
    \caption{Schematic representation of the experimental test rig, indicating the main geometric dimensions in $[\mathrm{mm}]$ and relative positions of its components.}
    \label{fig:rig_dimensions}
\end{figure}

The main geometric dimensions and component layout of the test rig are illustrated in Fig.~\ref{fig:rig_dimensions}, providing a schematic representation of the shaft length, bearing positions, and flange location.

The dataset files follow a standardized naming convention in the format \texttt{test\_<unbalance configuration>\_<rotation>}, in which each term explicitly describes the physical test conditions. The \texttt{<unbalance configuration>} field identifies the specific unbalance level applied to the system, while \texttt{<rotation>} denotes the operating speed of the main shaft. This convention ensures experimental traceability and allows each file to be unequivocally associated with the dynamic conditions under which the vibration signals were acquired. Additional details are provided in the repository README file.

\begin{figure}[H]
\centering
\includegraphics[width=0.5\linewidth]{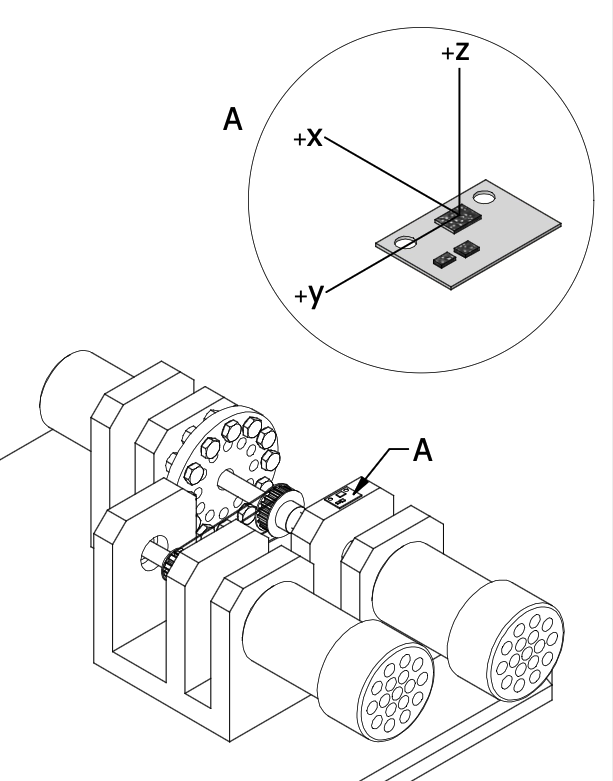}
\caption{Orientation of the triaxial accelerometer with respect to the experimental test rig, showing the X, Y, and Z axes used for vibration measurements.}
\label{fig:accelerometer_axes}
\end{figure}

All files are stored in CSV format using a comma delimiter and contain synchronized time series of triaxial vibration signals and shaft rotational speed. The header includes the columns \texttt{X}, \texttt{Y}, \texttt{Z}, \texttt{RPM}, \texttt{sec}, and \texttt{usec}. The variables \texttt{X}, \texttt{Y}, and \texttt{Z} correspond to the acceleration components measured along the three axes of the accelerometer, expressed in m/s$^2$, whose orientation with respect to the experimental test rig is illustrated in Fig.~\ref{fig:accelerometer_axes}, while \texttt{RPM} represents the instantaneous rotational speed estimated from the incremental encoder. The \texttt{sec} and \texttt{usec} columns store raw acquisition timestamps (seconds and microseconds), enabling precise reconstruction of the time base and ensuring temporal alignment between vibration measurements and operating conditions. A detailed description of each variable is provided in Table~\ref{tab:dataset}.

\begin{table}[h]
\caption{Structure of the dataset files, including variable names, units, data types, and corresponding physical quantities.}
\label{tab:dataset}
\begin{tabular}{@{}llll@{}}
\toprule
Parameter & Unit & Data Type & Description \\
\midrule

X & m/s² & Float64 & Acceleration along the X-axis \\

Y & m/s² & Float64 & Acceleration along the Y-axis \\

Z & m/s² & Float64 & Acceleration along the Z-axis \\

rpm & RPM & Float64 & Main shaft rotation \\

sec & s & int32 & Seconds (epoch format) \\

usec & $\mu s$ & int32 & Microseconds \\
\botrule
\end{tabular}
\end{table}

The acquisition parameters are summarized as follows: sampling frequency of 1~kHz; accelerometer configured to a $\pm$2~g measurement range; simultaneous acquisition along the three axes; incremental encoder with one pulse per 
revolution for rotational monitoring; and acquisition duration of 10 minutes per test to ensure steady-state operation.




\end{appendices}


\bibliography{sn-bibliography}

\end{document}